%% file: egpaper_for_review.tex
\documentclass[10pt,twocolumn,letterpaper]{article}

\usepackage{iccv}
\usepackage{times}
\usepackage{epsfig}
\usepackage{graphicx}
\usepackage{amsmath}
\usepackage{amssymb}
\usepackage{soul}

\usepackage[table]{xcolor}
\definecolor{mydarkblue}{rgb}{0,0.08,0.45}

\usepackage{algorithm}
\usepackage{algorithmic}
\usepackage{bbding}
\usepackage{booktabs}

\usepackage{multirow}
\usepackage{arydshln}
\usepackage[accsupp]{axessibility}
\usepackage{xcolor}
\usepackage{amsfonts}

\usepackage{comment}
\usepackage{color}
\usepackage{tabularx,ragged2e}
\usepackage{csquotes}
\usepackage{arydshln}
\usepackage{cite}
\usepackage{subfiles}
\usepackage{fontawesome}
\usepackage{inconsolata}
\usepackage{epsfig}
\usepackage{amsmath}
\usepackage{amssymb}
\usepackage{multirow}
\usepackage{booktabs}
\usepackage{cellspace}
\usepackage{subcaption}
\usepackage[hang,flushmargin]{footmisc}
\usepackage{booktabs,tabularx, colortbl} 
\usepackage{collcell}
\usepackage{pgfplots}
\pgfplotsset{compat=newest}
\usepgfplotslibrary{colormaps}
\usepackage{caption}
\usepackage{array}
\usepackage{enumitem}
\usepackage{adjustbox}

\definecolor{heatmap-yellow}{rgb}{1.0, 1.0, 0.8509803921568627}
\definecolor{heatmap-blue}{rgb}{0.03137254901960784, 0.11372549019607843, 0.34509803921568627}
\newcommand*\colourcheck[1]{%
  \expandafter\newcommand\csname #1check\endcsname{\textcolor{#1}{\ding{51}}}%
}
\newcommand*\colourcross[1]{%
  \expandafter\newcommand\csname #1cross\endcsname{\textcolor{#1}{\ding{55}}}%
}
\colourcheck{blue}
\colourcheck{green}
\colourcross{red}

\definecolor{audiogreen}{RGB}{234,241,230}
\definecolor{visualblue}{RGB}{231,240,244}

\definecolor{darkgreen}{rgb}{0.0, 0.75, 0.0}

\definecolor{bananayellow}{rgb}{1.0, 0.88, 0.21}

\usepackage{multirow}

\definecolor{lowcolor}{HTML}{ef8a62}
\definecolor{centralcolor}{HTML}{f7f7f7}
\definecolor{highcolor}{HTML}{67a9cf}
\definecolor{same_attr}{HTML}{009900}
\definecolor{diff_attr}{HTML}{FE0000}
\definecolor{Gray}{gray}{0.9}
\usepackage{color, colortbl}

\pgfplotsset{%
  colormap={PiYG}{%
    rgb=(0.93725490196, 0.54117647058, 0.38431372549)%
    rgb=(0.96862745098, 0.96862745098, 0.96862745098)%
    rgb=(0.40392156862, 0.66274509803, 0.81176470588)%
  }%
}

\def\pgfplotsshowcolormap#1{%
    \pgfplotscolormapifdefined{#1}{\relax}{%
        \pgfplotsset{colormap/#1}%
    }%
    \pgfplotscolormaptoshadingspec{#1}{1cm}\result
    \def\tempb{\pgfdeclarehorizontalshading{tempshading}{0.2cm}}%
    \expandafter\tempb\expandafter{\result}%
    \pgfuseshading{tempshading}%
}

\newlength{\savedtabcolsep}
\newcommand{\midsepremove}{\aboverulesep=0mm \belowrulesep=0mm}%
\usepackage{listofitems}
\setsepchar{ }

\newcommand{\ApplyGradient}[1]{%
    \readlist*\mylist{#1}%
  \pgfmathsetmacro{\Val}{\mylist[1]}
  \pgfmathsetmacro{\MiddleVal}{\MinVal + (\MaxVal - \MinVal) / 2}%
  \ifdim \Val pt > \MiddleVal pt%
      \pgfmathsetmacro{\PercentColor}{max(min(100.0*(\Val - \MiddleVal)/(\MaxVal-\MiddleVal),100.0),0.00)}%
      \edef\HeatCell{\noexpand\cellcolor{highcolor!\PercentColor!centralcolor}}%
      \HeatCell$#1$%
  \else
      \pgfmathsetmacro{\PercentColor}{max(min(100.0*(\MiddleVal - \Val)/(\MiddleVal-\MinVal),100.0),0.00)}%
      \edef\HeatCell{\noexpand\cellcolor{lowcolor!\PercentColor!centralcolor}}%
      \HeatCell$#1$%
  \fi%
}

\newcommand{\ApplyGradientReverse}[1]{%
    \readlist*\mylist{#1}%
  \pgfmathsetmacro{\Val}{\mylist[1]}%
  \pgfmathsetmacro{\MiddleVal}{\MinVal + (\MaxVal - \MinVal) / 2}%
  \ifdim \Val pt > \MiddleVal pt%
      \pgfmathsetmacro{\PercentColor}{max(min(100.0*(\Val - \MiddleVal)/(\MaxVal-\MiddleVal),100.0),0.00)}%
      \edef\HeatCell{\noexpand\cellcolor{lowcolor!\PercentColor!centralcolor}}%
      \HeatCell$#1$%
  \else
      \pgfmathsetmacro{\PercentColor}{max(min(100.0*(\MiddleVal - \Val)/(\MiddleVal-\MinVal),100.0),0.00)} %
      \edef\HeatCell{\noexpand\cellcolor{highcolor!\PercentColor!centralcolor}}%
      \HeatCell$#1$%
  \fi%
}

\newcolumntype{\C}[2]{>{\def\MinVal{#1}\def\MaxVal{#2}\collectcell\ApplyGradient}c<{\endcollectcell}}

\newcolumntype{\CR}[2]{>{\def\MinVal{#1}\def\MaxVal{#2}\collectcell\ApplyGradientReverse}c<{\endcollectcell}}

\usepackage[many]{tcolorbox}
\newtcolorbox{myboxi}[2][]{
  breakable,
  title=#1,
  colback=white,
  colbacktitle=white,
  coltitle=black,
  fonttitle=\itshape,
  bottomrule=-0.1pt,
  toprule=-0.2pt,
  leftrule=5pt,
  rightrule=0pt,
  titlerule=0pt,
  arc=0pt,
  boxsep=1mm,
  outer arc=0pt,
  colframe=#2,
}

\usepackage[pagebackref=true,breaklinks=true,letterpaper=true,colorlinks,bookmarks=false]{hyperref}

\iccvfinalcopy % *** Uncomment this line for the final submission

\ificcvfinal\fi

\begin{document}

%%%%%%%%% TITLE
%\title{Tubelet contrastive video representation learning enables data-efficiency and better generalization}
\title{Tubelet-Contrastive Self-Supervision for Video-Efficient Generalization}
%\title{Better generalization, data efficiency  with Tubelet contrastive video representation learning }

\author{First Author\\
Institution1\\
Institution1 address\\
{\tt\small firstauthor@i1.org}
\and
Second Author\\
Institution2\\
First line of institution2 address\\
{\tt\small secondauthor@i2.org}
}

\author{Fida Mohammad Thoker, Hazel Doughty, Cees  G. M.  Snoek\\
University of Amsterdam}

\maketitle
%\ificcvfinal\thispagestyle{empty}\fi

% Commented our for Appendix submission
\input{sections/0_abstract}
\input{sections/1_introduction}

\input{sections/2_related_work}

\input{sections/3_method}

\input{sections/4_expirements}

\input{sections/5_expirements}

\input{sections/6_expirements}

\input{sections/7_conclusion}

{\small
\bibliographystyle{ieee_fullname}
\bibliography{egbib}
}
\appendix
\input{sections/appendix}

\end{document}

%% file: sections/0_abstract.tex
\begin{abstract}
\vspace{-0.8em}
We propose a self-supervised method for learning motion-focused video representations. Existing approaches minimize distances between temporally augmented videos, which maintain high spatial similarity. We instead propose to learn similarities between videos with identical local motion dynamics but an otherwise different appearance. We do so by adding synthetic motion trajectories to videos which we refer to as tubelets. 
By simulating different tubelet motions and applying transformations, such as scaling and rotation, we introduce motion patterns beyond what is present in the pretraining data. This allows us to learn a video representation that is remarkably data efficient: our approach maintains performance when using only 25\% of the pretraining videos. Experiments on 10 diverse downstream settings 
demonstrate our competitive performance and generalizability to new domains and fine-grained actions. Code is available at \href{https://github.com/fmthoker/tubelet-contrast}{https://github.com/fmthoker/tubelet-contrast}.
\end{abstract}

%

%% file: sections/1_introduction.tex
%!TEX root = egpaper_for_review.tex
\vspace{-3em}
\section{Introduction}
\begin{figure}[t!]
\centering
\includegraphics[width=0.95\linewidth]{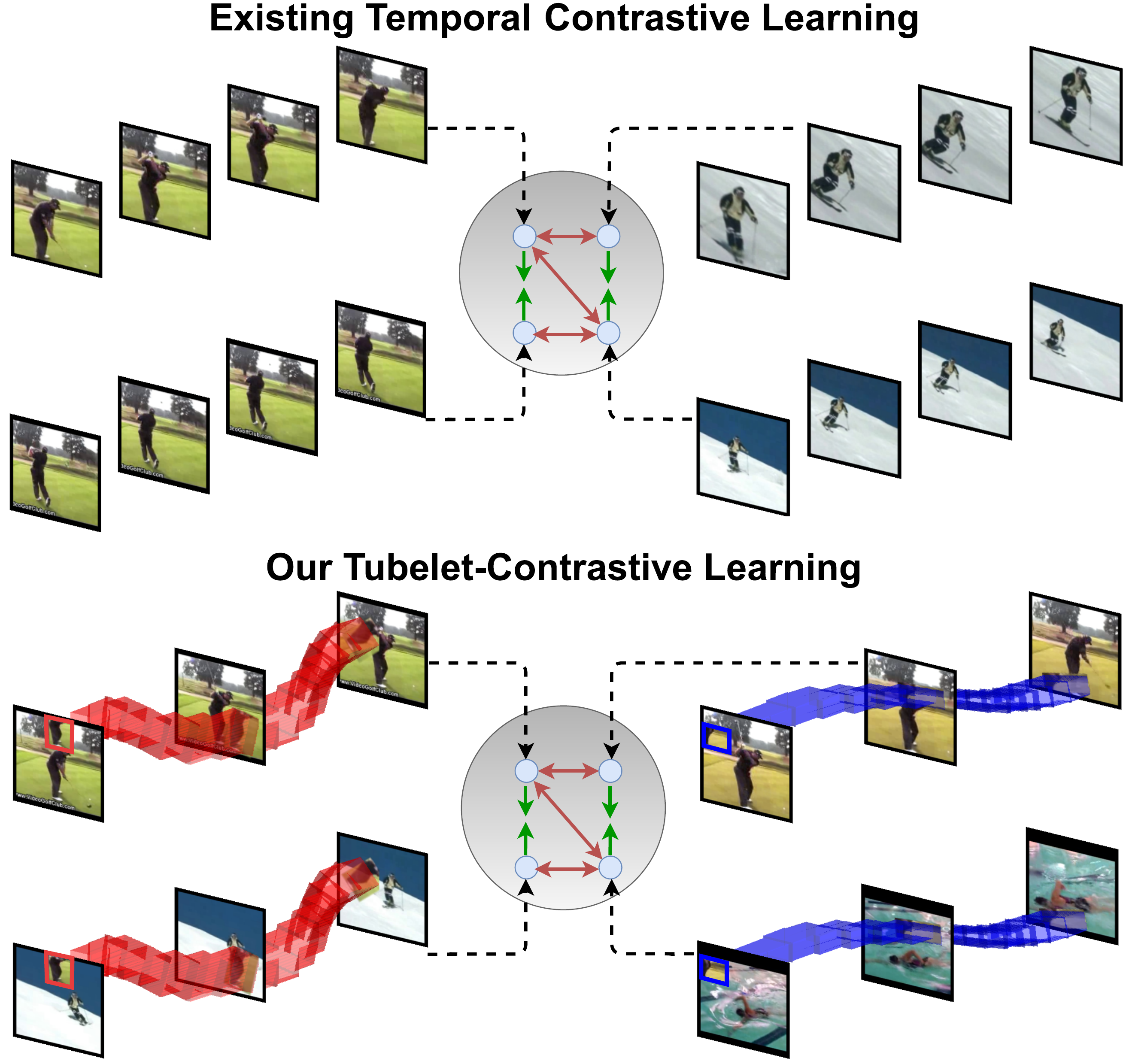}
\vspace{-0.8em}
\caption{\textbf{Tubelet-Contrastive Positive Pairs} (bottom) only share the spatiotemporal motion dynamics inside the simulated tubelets, while temporal contrastive pairs (top) suffer from a high spatial bias. Contrasting tubelets results in a data-efficient and generalizable video representation. 
}
\label{fig:clips}
\vspace{-0.8em}
\end{figure}

\vspace{-0.3em}
This paper aims to learn self-supervised video representations, useful for distinguishing actions. In a community effort to reduce the manual, expensive, and hard-to-scale annotations needed for many downstream deployment settings, the topic has witnessed tremendous progress in recent years~\cite{odd,xu2019self,simon, schiappa2022self}, particularly through contrastive learning~\cite{cvrl,videomoco-pan2021videomoco,rspnet-chen2020RSPNet,large-scale-feichtenhofer2021large}. Contrastive approaches learn representations through instance discrimination~\cite{infonce}, by increasing feature similarity between spatially and temporally augmented clips from the same video. Despite temporal differences, such positive video pairs often maintain high spatial similarity (see Figure~\ref{fig:clips}), allowing the contrastive task to be solved by coarse-grained features without explicitly capturing local motion dynamics. This limits the generalizability of the learned video representations, as shown in our prior work~\cite{thoker2022severe}. Furthermore, prior approaches are constrained by the amount and types of motions present in the pretraining data. This makes them data-hungry, as video data has high redundancy with periods of little to no motion. In this work, we address the need for data-efficient and generalizable self-supervised video representations by proposing a contrastive method to learn local motion dynamics.

We take inspiration from action detection, where tubelets are used to represent the motions of people and objects in videos through bounding box sequences %commonly used to represent the sequence of bounding boxes encompassing actors and their actions as they appear and move throughout a video
\eg,\cite{jain2014action,kalogeiton2017action,li2020actions}. Typically, many tubelet proposals are generated for a video, which are processed to find the best prediction. %fitting one for the classification objective of interest. 
Rather than finding tubelets in video data, we simulate them. %, similar to seminal work on learning optical flow~\cite{dosovitskiy2015flownet}. 
In particular, we sample an image patch and `paste' it with a randomized motion onto two different video clips as a shared tubelet (see Figure~\ref{fig:clips}). These two clips form a positive pair for contrastive learning where the model has to rely on the spatiotemporal dynamics of the tubelet to learn the similarity. With such a formulation, we can simulate a large variety of motion patterns that are not present in the original videos. This allows our model to be data-efficient while improving generalization to new domains and fine-grained actions.

We make four contributions. First, we explicitly learn from local motion dynamics in the form of synthetic tubelets and design a simple but effective tubelet-contrastive framework. Second, we propose different ways of simulating tubelet motion and transformations to generate a variety of motion patterns for learning. Third, we reveal the remarkable data efficiency of our proposal: on five action recognition datasets our approach maintains performance when using only 25\% of the pretraining videos. What is more, with only 5-10\% of the videos we still outperform the vanilla contrastive baseline with 100\% pretraining data for several datasets. Fourth, our comparative experiments on 10 downstream settings, including UCF101~\cite{UCF-101-arxiv}, HMDB51~\cite{HMDB-51-ICCV}, 
Something Something~\cite{SS-v2-arxiv}, and FineGym~\cite{Gym-99-arxiv}, %and the complete SEVERE Benchmark~\cite{thoker2022severe}, 
further demonstrate our competitive performance, generalizability to new domains, and suitability of our learned representation for fine-grained actions.

%% file: sections/2_related_work.tex
\vspace{-0.3em}
\section{Related Work}
\vspace{-0.3em}
\noindent\textbf{Self-Supervised Video Representation Learning.} 
The success of contrastive learning in images~\cite{moco, chen2020simple,grill2020bootstrap,moskalev2022contrasting} inspired many video contrastive works~\cite{cvrl,videomoco-pan2021videomoco,rspnet-chen2020RSPNet,huang2021ascnet,srtc,pretext-contrast}. Alongside spatial invariances, these works learn invariances to temporal crops~\cite{videomoco-pan2021videomoco,cvrl,svt} and video speed~\cite{rspnet-chen2020RSPNet,huang2021ascnet,srtc}. Some diverge from temporal invariances and encourage equivariance~\cite{gdt-patrick2020multimodal,dave2022tclr} to learn finer temporal representations. 
For instance, TCLR~\cite{dave2022tclr} enforces within-instance temporal feature variation, while TE~\cite{jenni2021time_eqv} learns equivariance to temporal crops and speed with contrastive learning. Alternatively, many works learn to predict temporal transformations such as clip order~\cite{misra2016shuffle,xu2019self,sorting,odd}, speed~\cite{benaim2020speednet, cho2020self,prp} and their combinations~\cite{vcp,simon}. These self-supervised temporal representations are effective for classifying and retrieving coarse-grained actions but are challenged by downstream settings with subtle motions \cite{thoker2022severe, schiappa2022self}.
Other works utilize the multimodal nature of videos~\cite{alwassel_2020_xdc,selavi,gdt-patrick2020multimodal,avid-cma-morgado2021audio,coclr,motion_fit,mac2019learning} and learn similarity with audio~\cite{selavi,alwassel_2020_xdc,avid-cma-morgado2021audio} and optical flow~\cite{coclr,motion_fit, mscl, xiao2022maclr}.  
We contrast motions of synthetic tubelets to learn a video representation from only RGB data that can generalize to tasks requiring fine-grained motion understanding.

Other self-supervised works learn from the spatiotemporal dynamics of video.
Both BE~\cite{background_removing} and FAME~\cite{fame} remove background bias by adding static frames~\cite{background_removing} or replacing the background~\cite{fame} in positive pairs. %Similarly, FAME~\cite{fame} merges the foreground from one video with the background from another video to generate positive pairs for contrastive learning to enhance motion awareness in the learned video representations. 
Several works instead use masked autoencoding to learn video representations~\cite{tong2022videomae, feichtenhofer2022masked}.
However, these works are all limited to the motions present in the pretraining dataset. We prefer to be less dataset-dependent and generate synthetic motion tubelets for contrastive learning, which also offers a considerable data-efficiency benefit.
CtP~\cite{ctp-wang2021unsupervised} and MoSI~\cite{motion_static} both aim to predict motions in pretraining. CtP~\cite{ctp-wang2021unsupervised} learns to track image patches in video clips while MoSI~\cite{motion_static} learns to predict the speed and direction of added pseudo-motions. We take inspiration from these works and contrast synthetic motions from tubelets which allows us to learn generalizable and data-efficient representations.
%
%Different from CtP and MoSI we design a tubelet based contrastive framework for learning similarities and dissimilarities between simulated motion patterns to enhance motion-focused video representation learning.
%Both BE and Fame focus on removing spatial basis in positive pairs but are limited by the motions patterns present in original video clips, while we simulate a variety of synthetic motion patterns for contrastive learning in the form of tubelet motions and tubelet transformations which enhances downstream generalization and data efficency.
%
\begin{figure*}[t!]
\centering
\vspace{-1em}
\includegraphics[width=0.95\linewidth]{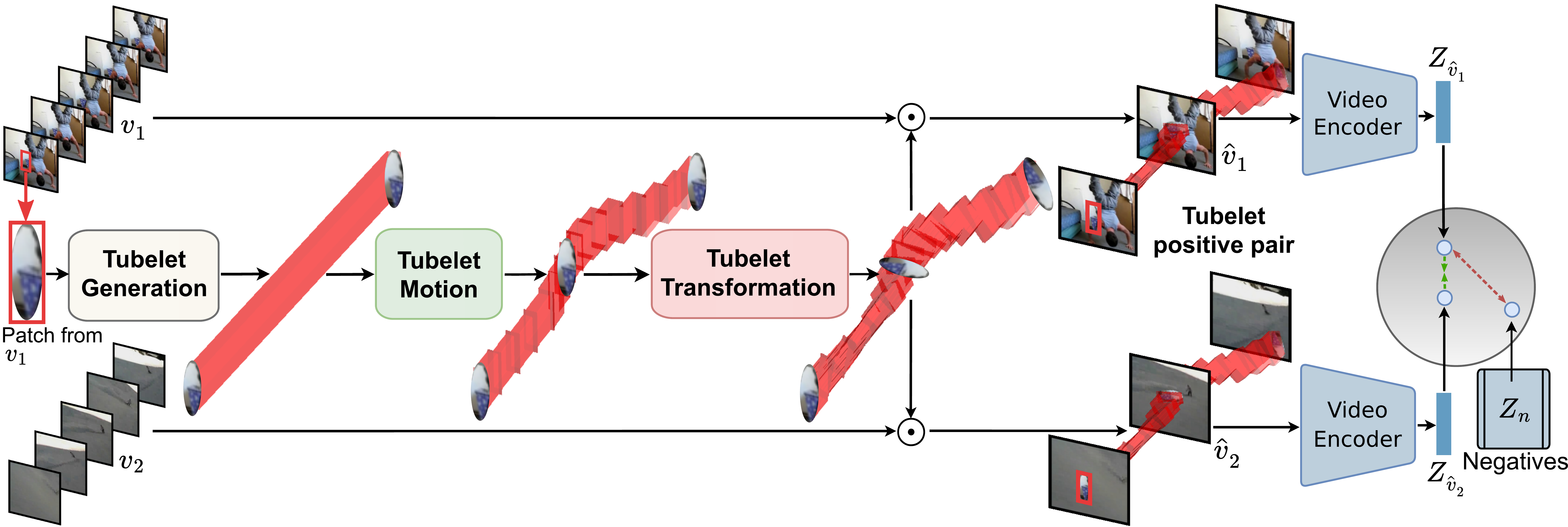}
\vspace{-1em}
\caption{\textbf{Tubelet-Contrastive Learning.} We sample two clips ($v_1, v_2$) from different videos and randomly crop an image patch from $v_1$. We generate a tubelet by replicating the patch in time and add motion through a sequence of target locations for the patch. 
We then add complexity to these motions by applying transformations, such as rotation, to the tubelet. The tubelet is overlaid $\odot$ onto both  clips to form a positive tubelet pair ($\hat{v}_{1}, \hat{v}_{2}$). We learn similarities between clips with the same tubelets (positive pairs) and dissimilarities between clips with different tubelets (negatives) using a contrastive loss.}
\label{fig:framework}
\vspace{-1.2em}
\end{figure*}

\noindent\textbf{Supervised Fine-Grained Motion Learning. } 
%  add MLFO and MSCl
%- introduce coarse grained vs fine-grained
%- fine-grained datasets
%- fine-grained needs motion
%- approaches for fine-grained include NN blocks for motion, decoupling spatial and motion features, and aggregating features at different temporal resolutions. Also optical flow.
%- tasks beyond action recognition that aim for motions, repetition counting, recognizing adverbs, others?
%- these are supervised, we aim for self-supervised representation learnt from coarse-grained and generalizable to fine-grained, new domains.
%
While self-supervised works have mainly focused on learning representations to distinguish coarse-grained actions, much progress has been made in supervised learning of motions. 
Approaches distinguish actions by motion-focused neural network blocks~\cite{kwon2021learning, mac2019learning, lin2019tsm, kim2021relational}, decoupling motion from appearance~\cite{rahmani2022dynamic, sun2022fine}, aggregating multiple temporal scales~\cite{yang2020temporal, ni2014multiple, feichtenhofer2019slowfast}, and sparse coding to obtain a mid-level motion representation~\cite{mavroudi2018end, piergiovanni2018fine, shao2020intra}. 
Other works exploit skeleton data~\cite{duan2022revisiting, hong2021video} or optical flow~\cite{simonyan2014two, feichtenhofer2016convolutional}. Alternatively, several works identify motion differences within an action class, by repetition counting~\cite{hu2022transrac, ucfrep-zhang2020context, zhang2021repetitive}, recognizing adverbs~\cite{doughty2020action, doughty2022you} or querying for action attributes~\cite{tqn}.  
Different from all these works, we learn a motion-sensitive video representation with self-supervision. We do so by relying on just coarse-grained video data in pretraining and demonstrate downstream generalization to fine-grained actions.

 \noindent\textbf{Tubelets. } 
 Jain \etal defined tubelets as class-agnostic sequences of bounding boxes over time~\cite{jain2014action}.  Tubelets can represent the movement of people and objects and are commonly used for object detection in videos\cite{kang2017object,kang2017t,feichtenhofer2017detect}, spatiotemporal action localization~\cite{kalogeiton2017action,yang2019step,li2020actions,jain2014action,hou2017tube,zhao2022tuber} and video relation detection~\cite{ChenRelationICCV21}. Initially, tubelets were obtained by supervoxel groupings and dense trajectories \cite{jain2014action,GemertBMVC15} and later from 2D CNNs~\cite{kalogeiton2017action,li2020actions}, 3D CNNs~\cite{hou2017tube,yang2019step} and transformers~\cite{zhao2022tuber}. We introduce (synthetic) tubelets of pseudo-objects for contrastive video self-supervised learning.

%% file: sections/3_method.tex
\vspace{-0.3em}
\section{Tubelet Contrast}
\vspace{-0.3em}
We aim to learn motion-focused video representations from RGB video data with self-supervision. After revisiting temporal contrastive learning, we propose tubelet-contrastive learning to reduce the spatial focus of video representations and instead learn similarities between spatio-temporal tubelet dynamics (Section~\ref{section:tbcl}). We encourage our representation to be motion-focused by simulating a variety of tubelet motions (Section~\ref{section:motions}). To further improve data efficiency and generalizability, we add complexity and variety to the motions through tubelet transformations (Section~\ref{section:transformations}). Figure~\ref{fig:framework} shows an overview of our approach.

%\subsection{Video-Contrastive Learning}
%\label{section:tcl}
\noindent\textbf{Temporal Contrastive Learning.} Temporal contrastive learning learns feature representations via instance discrimination \cite{infonce}. This is achieved by maximizing the similarity between augmented clips from the same video (positive pairs) and minimizing the similarity between clips from different videos (negatives). Concretely given a set of videos $V$, the positive pairs $(v,v')$ are obtained by sampling different temporal crops of the same video~\cite{videomoco-pan2021videomoco,rspnet-chen2020RSPNet} and applying spatial augmentations such as cropping and color jittering. Clips sampled from other videos in the training set act as negatives. The extracted clips are passed through a video encoder and projected on a representation space by a non-linear projection head to obtain clip embeddings $(Z_{v},Z_{v'})$. The noise contrastive estimation loss InfoNCE \cite{infonce} is used for the optimization: 
%%%%%%% %%%%%%%%%%%
\vspace{-0.7em}
\begin{equation}
\resizebox{0.9\linewidth}{!}{
$\mathcal{L}_{contrast}{(v, v')} = -\log \frac{\displaystyle h(Z_{v},Z_{v'})}
{ \displaystyle h(Z_{v},Z_{v'})+\displaystyle \sum_{Z_n \sim \mathcal{N}} h(Z_{v},Z_{n})}$
}
\label{eq:tcl}
\end{equation}
where $h(Z_{v},Z_{v'}) {=} \mathrm{exp}(Z_{v}\cdot Z_{v'}/\tau)$, $\tau$ is the temperature parameter and $\mathcal{N}$ is a set of negative clip embeddings. %\hd{Add sentence on why this loss/way of contrasting is insufficient for us.}

\vspace{-0.3em}
\subsection{{Tubelet-Contrastive Learning}}
\vspace{-0.3em}
\label{section:tbcl}
Different from existing video contrastive self-supervised methods, we explicitly aim to learn motion-focused video representations while relying only on RGB data. To achieve this we propose to learn similarities between simulated tubelets.  Concretely, we first generate tubelets in the form of moving patches which are then overlaid onto two different video clips to generate positive pairs that have a high motion similarity and a low spatial similarity. Such positive pairs are then employed to learn video representations via instance discrimination, allowing us to learn more generalizable and motion-sensitive video representations.

\noindent\textbf{Tubelet Generation.} 
We define a tubelet as a sequence of object locations in each frame of a video clip. Let's assume an object $p$ of size ${H}'\times{W}'$ moving in a video clip $v$ of length $T$. Then the tubelet is defined as follows:
\vspace{-0.3em}
\begin{equation}
\mathrm{Tubelet}_{p} = [(x^1,y^1) ,..,(x^T,y^T)],
\label{eq:tubelet_def}
\vspace{-0.5em}
\end{equation}
 where $(x^i,y^i)$  is the center coordinate of the object $p$ in frame $i$ of clip $v$. % such that $0<x^i<W$ and $<0<y^i<H$, $W$ and $H$ is is width and height of the clip $v$. 
 For this work, a random image patch of size ${H}'\times{W}'$ acts as a pseudo-object overlaid on a video clip to form a tubelet. To generate the tubelet we first make the object appear static, \ie, $x^1{=}x^2{=}...{=}x^T$ and $y^1{=}y^2 = ...{=}y^T$, and explain how we add motion in Section~\ref{section:motions}. % we explain how we define the tubelet to give the pseudo-object motion.
 
\noindent\textbf{Tubelet-Contrastive Pairs.} To create contrastive tubelet pairs, we first randomly sample clips $v_1$ and $v_2$ of size $H{\times}W$ and length $T$ from two different videos in $V$. From $v_1$ we randomly crop an image patch $p$ of size ${H}' \times {W}'$.  such that ${H}' \ll H$  and  ${W}' \ll W$.  
From the patch $p$, we  construct a tubelet $\mathrm{Tubelet}_{p}$ as in Eq.~\eqref{eq:tubelet_def}.
Then, we overlay the generated tubelet $\mathrm{Tubelet}_{p}$ onto both $v_1$ and $v_2$ to create two modified video clips $\hat{v}_{1}$ and $\hat{v}_{2}$: %\cs{There is a lot of white space around equations, probably caused by split command, pls fix.}
%\vspace{-0.6em}
\abovedisplayskip=4pt
\belowdisplayskip=4pt
\begin{align}
 &\hat{v}_{1} = v_{1} \odot \mathrm{Tubelet}_{p} \quad
 &\hat{v}_{2} = v_{2} \odot \mathrm{Tubelet}_{p},
\label{eq:overlay_single_tubelet}
\end{align}
where $\odot$ refers to pasting patch $p$ in each video frame at locations determined by $\mathrm{Tubelet}_{p}$. Eq.~\eqref{eq:overlay_single_tubelet} can be extended for a set of $M$ tubelets $\{\mathrm{Tubelet}_{p_1},...,\mathrm{Tubelet}_{p_M}\}$ from $M$ patches randomly cropped from $v_1$ as:
\begin{align}
 \hat{v}_{1} = v_{1} \odot \{\mathrm{Tubelet}_{p_1},...,\mathrm{Tubelet}_{p_M}\} \nonumber \\
 \hat{v}_{2} = v_{2} \odot \{\mathrm{Tubelet}_{p_1},...,\mathrm{Tubelet}_{p_M}\}.
\label{eq:overlay_M_tubelets}
\end{align}
As a result, $\hat{v}_{1}$ and $\hat{v}_{2}$ share the spatiotemporal dynamics of the moving patches in the form of tubelets and have low spatial bias since the two clips come from different videos.  %\cs{Figure~\ref{fig:clips} shows examples of $v_{1}'$ and $v_{2}'$.}
Finally, we adapt the contrastive loss from Eq.~\eqref{eq:tcl} and apply $\mathcal{L}_{contrast}(\hat{v}_1, \hat{v}_2)$. %from Equation \eqref{eq:tcl} to explicitly contrast tubelet pairs as follows: \cs{Is this equation really needed?}
% \begin{equation}
%\mathcal{L}_{\cs{TBCL}}{(V_1,V_2)} = -\log \frac{\displaystyle h(Z_{v_{1}'}, Z_{v_{2}'})}
%{ \displaystyle h(Z_{v_{1}'}, Z_{v_{2}'}) + \displaystyle\sum_{Z_{v_{n}'} \sim \mathcal{N}} h(Z_{v_{1}'}, Z_{v_{n}'}) }
%\label{eq:tubelet_contrastive_loss}
%\end{equation}
%where  $Z_{v_{1}'},Z_{v_{2}'}$ are the embeddings of the positive tubelet pair ($v_{1}',v_{2}'$),  $h(Z_{v_{1}'},Z_{v_{2}'}) {=} \mathrm{exp}(Z_{v_{1}'}\cdot Z_{v_{2}'}/\tau)$, $\tau$ is the temperature softening hyper-parameter and 
Here the set of negatives $\mathcal{N}$ contains videos with different tubelets. Since the only similarity in positive pairs is the tubelets, the network must rely on temporal cues causing a motion-focused video representation. 

\vspace{-0.3em}
\subsection{Tubelet Motion}
\vspace{-0.3em}
\label{section:motions}
%As discussed in the previous section, our aim is to learn the same feature representations for video clips that have the same motion tubelets and different feature representations for the clips with different motion tubelets. Thus, in order 
%To learn good video features, which are able to represent a variety of motion patterns, we need to simulate a variety of tubelet trajectories for learning. 
To learn motion-focused video representations, we need to give our tubelets motion variety. Here, we discuss how to simulate motions by generating different patch movements in the tubelets. Recall, Eq.~\eqref{eq:tubelet_def} defines a tubelet by image patch $p$ and its center coordinate in each video frame. We consider two types of tubelet motion: linear and non-linear. 

\begin{figure}[t!]
\centering
\includegraphics[width=\linewidth]{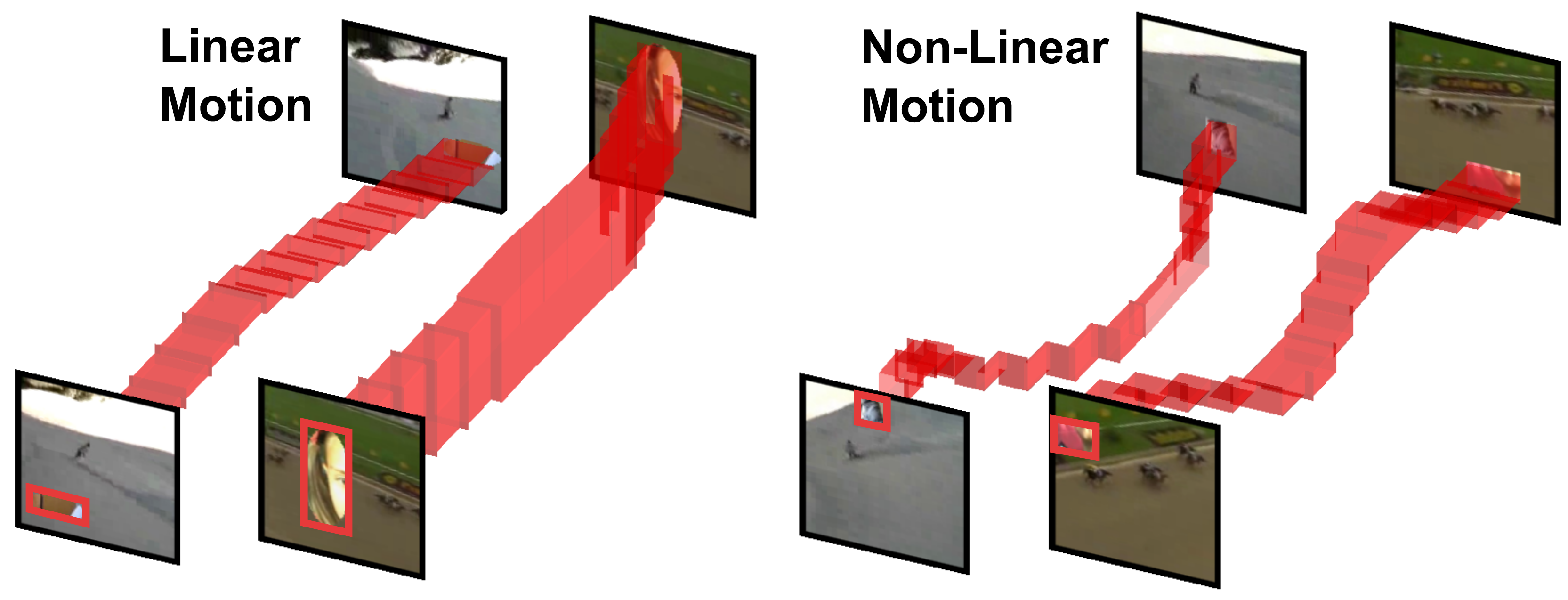}
\vspace{-2.3em}
\caption{ \textbf{Tubelet Motion.} Examples for \textit{Linear} (left) and \textit{Non-Linear} (right). Non-linear motions enable the simulation of a larger variety of motion patterns to learn from. 
}
\label{fig:traj}
\vspace{-1.3em}
\end{figure}

\noindent\textbf{Linear Motion.} %First, we aim to simulate motions where the patches move along a linear path.  
We randomly sample the center locations for the patch in $K$ keyframes: the first frame ($i{=}1$), the last frame ($i{=}T$),  and $K{-}2$ randomly selected frames. These patch locations are sampled from uniform distributions $x\in[0,W]$ and $y\in[0,H]$, where $W$ and $H$ are the video width and height. 
%To ensure smoothness, we constrain the difference between the center locations in neighboring keyframes to be less than $\Delta$ pixels.
%as $(x^1,y^1)$, $(x^{k1},y^{k1}),..,(x^{K-2},y^{K-2})$ and $(x^T,y^T)$.
Patch locations for the remaining frames $i\notin K$ are then linearly interpolated between keyframes so we obtain the following linear motion definition:
%\begin{equation}
\begin{align}
\label{eq:linear_motion}
%\quad 
&  \mathrm{Tubelet}^{\mathrm{Lin}} = [(x^1,y^1),(x^2,y^2),...,(x^T,y^T)], \text{ s.t.} \\ 
& (x^i,y^i) = 
\begin{cases}
(\mathcal{U}(0,W),\mathcal{U}(0,H)), & \text{if } i \in K\\
\mathrm{Interp}((x^{k},y^{k}),(x^{k+1},y^{k+1})), & \text{otherwise} \nonumber
\end{cases}
\end{align}
%The center locations for the remaining frames $i\neq(1,m,T)$ are then linearly interpolated between the neighboring keyframes.  
where $\mathcal{U}$ is a function for uniform sampling, $k$ and $k{+}1$ are the neighboring keyframes to frame $i$ and $\mathrm{Interp}$ gives a linear interpolation between keyframes. %and $(x_{i}^{k},y_{i}^{k}),(x_{i}^{k+1},y_{i}^{k+1})$ are the center coordinates in these keyframes. 
To ensure smoothness, we constrain the difference between the center locations in neighboring keyframes to be less than $\Delta$ pixels.
This formulation results in tubelet motions where patches follow linear paths across the video frames. %As a result, we generate motion patterns for learning that simulate objects with linear movements in real videos. 
%where $\Delta i$ is the index difference between the two keyframes. 
%\cs{I do not like this visualization, pls update as discussed}
The left of Figure~\ref{fig:traj} shows examples of  such linear tubelet motions.

\noindent \textbf{Non-Linear Motion.} 
Linear motions are simple and limit the variety of motion patterns that can be generated.  Next, we simulate motions where patches move along more complex non-linear paths, to better emulate motions in real videos. %In contrast to linear movements, such paths generate a variety of more realistic motion patterns to learn from. 
We create non-linear motions by first sampling $N$ 2D coordinates ($N \gg T$) uniformly from $x \in [0,W]$ and $y \in [0,H]$.  Then, we apply a $1D$ Gaussian filter along $x$ and $y$ axes to generate a random smooth nonlinear path as:%\cs{sloppy equation, pls fix}
\vspace{-1em}
\begin{align}
\label{eq:nonlinear_motion}
\quad 
\mathrm{Tubelet}^{\mathrm{NonLin}} = &[(g(x^1),g(y^1)),...,(g(x^N),g(y^N))] \nonumber \\ 
\text{s.t. } \quad
&g(z) = \frac{1}{\sqrt[]{2 \pi } \sigma} e^{- {z}^{2}/{2 \sigma ^{2}}} %\ \text{where} \ x^i\in[0,W]\nonumber \\
%&y^i \in [0,H], \hspace{5pt} 
%&g(y^i) = \frac{1}{\sqrt[]{2 \pi } \sigma} e^{- {y^i}^{2}/{2 \sigma ^{2}}},
%\text{where} \ y^i\in[0,H] 
\end{align}
where $\sigma$ is the smoothing factor for the gaussian kernels. %and $W,H$ is the width and height of the video clips. 
Note the importance of sampling $N\gg T$ points to ensure a non-linear path. If $N$ is too small then the  path becomes linear after gaussian smoothing. %We also ensure that tubelets are sampled at different XY positions in the video clip and have varying lengths of pixel movement along the time axis.
%random scaling and translation is applied on the $\mathrm{Tubelet}^{\mathrm{NonLin}}$ generated from Equation \eqref{eq:nonlinear_motion} as:
%Moreover, to ensure that tubelets are sampled at different XY positions in the video clip and have varying lengths of pixel movement, random scaling and translation is applied on the $\mathrm{Tubelet}^{\mathrm{NonLin}}$ generated from Equation \eqref{eq:nonlinear_motion} as:
%\begin{equation}
%\mathrm{Tubelet}^{\mathrm{Scale}}_\mathrm{{Translate}} =  a *  \mathrm{Tubelet}^{\mathrm{NonLin}} + c, \\ 
%\label{eq:scale_transaltion}
%\end{equation}
%
 %where $a{=}(S_x,S_y)$ are the scaling factors and $c{=}(T_x,T_y)$  are translation factors along $X$ and $Y$ coordinates and are randomly sampled such that $0 < S_x < W$, $0 < S_y < H$, $0 < T_x < W$ and $0 < T_y < H$.
 %Note that the same scaling and translation  is applied to each $2D$ point in $\mathrm{Tubelet}^{\mathrm{NonLin}}$. 
 We downsample the resulting non-linear tubelet in Eq.~\eqref{eq:nonlinear_motion} from $N$ to $T$ coordinates resulting in the locations for patch $p$ in the $T$ frames.
The right of Figure~\ref{fig:traj}
shows examples of non-linear tubelet motions.

\begin{figure}[t!]
\centering
\includegraphics[width=\linewidth]{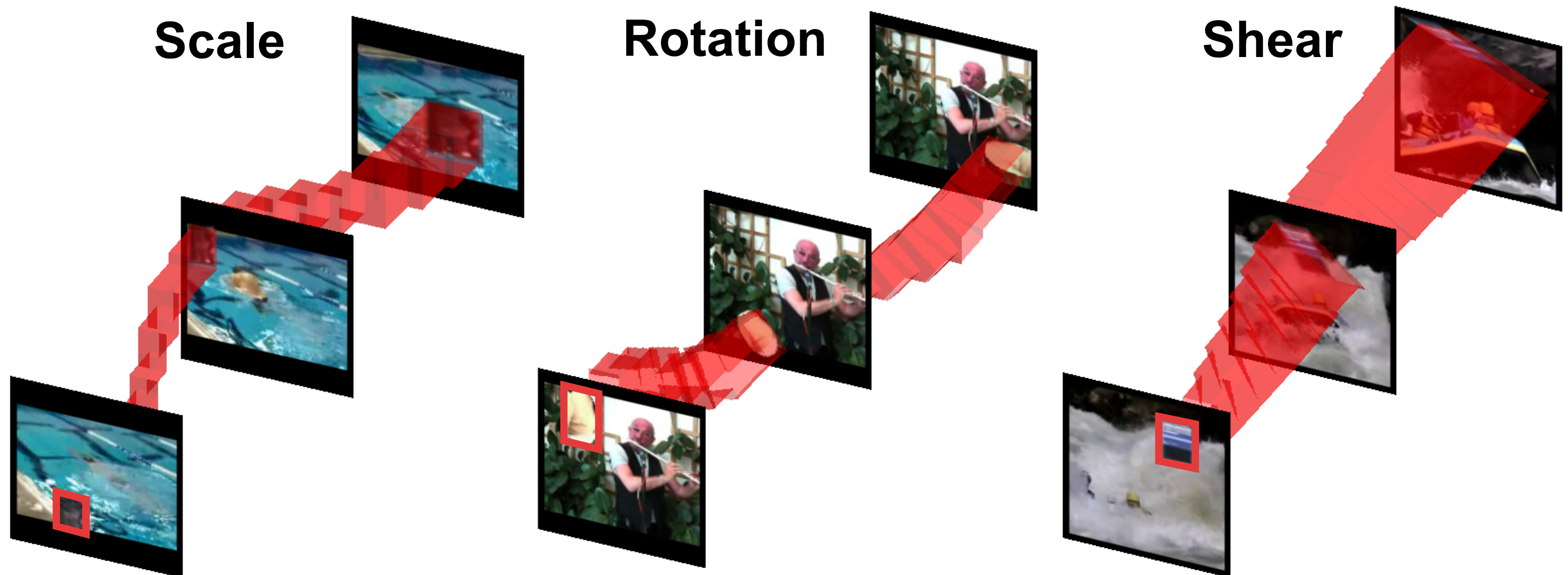}
\vspace{-2em}
\caption{\textbf{Tubelet Transformation.} Examples for
\textit{Scale} (left), \textit{Rotation} (middle), and \textit{Shear} (right). The patch is transformed as it moves through the tubelet.  
}
\label{fig:transformations}
\vspace{-1em}
\end{figure}

\begin{table*}[t]
    \centering
    \resizebox{\linewidth}{!}{
    \begin{tabular}{llllrrrl}
    \toprule
    \textbf{Evaluation Factor} & \textbf{Experiment} & \textbf{Dataset} & \textbf{Task} & \textbf{\#Classes} & \textbf{\#Finetuning} & \textbf{\#Testing} & \textbf{Eval Metric}\\
    \midrule

    \multirow{2}{*}{\textbf{Standard}} & UCF101 & UCF 101~\cite{UCF-101-arxiv} & Action Recognition & 101 & 9,537 & 3,783 & Top-1 Accuracy\\
 & HMDB51 & HMDB 51~\cite{HMDB-51-ICCV} & Action Recognition & 51 & 3,570 & 1,530 & Top-1 Accuracy\\
    \midrule
    \multirow{2}{*}{\textbf{Domain Shift}} & SSv2 & Something-Something~\cite{SS-v2-arxiv} & Action Recognition & 174 & 168,913 &  24,777 & Top-1 Accuracy\\
     & Gym99 & FineGym~\cite{Gym-99-arxiv} & Action Recognition & 99 & 20,484 & 8,521 & Top-1 Accuracy\\
    \midrule
    \multirow{2}{*}{\textbf{Sample Efficiency}}& UCF ($10^3$) & UCF 101~\cite{UCF-101-arxiv} & Action Recognition & 101 & 1,000 & 3,783 & Top-1 Accuracy\\
     & Gym ($10^3$) & FineGym~\cite{Gym-99-arxiv} & Action Recognition & 99 & 1,000 & 8,521 & Top-1 Accuracy\\
    \midrule
    \multirow{2}{*}{\textbf{Action Granularity}} & FX-S1 & FineGym~\cite{Gym-99-arxiv} & Action Recognition & 11 & 1,882 & 777 & Mean Class Acc\\
     & UB-S1 & FineGym~\cite{Gym-99-arxiv} & Action Recognition & 15 & 3,511 & 1,471 & Mean Class Acc\\
    \midrule
    \multirow{2}{*}{\textbf{Task Shift}} & UCF-RC & UCFRep~\cite{ucfrep-zhang2020context} & Repetition Counting & - & 421 & 105 & Mean Error\\
     & Charades & Charades~\cite{charades-sigurdsson:hal-01418216} & Multi-label Recognition & 157 & 7,985 & 1,863 & mAP\\
    \bottomrule
    \end{tabular}}
    \vspace{-0.8em}
    \caption{\textbf{Benchmark Details} for the  downstream evaluation factors, experiments, and datasets we cover. For non-standard evaluations, we follow the SEVERE benchmark~\cite{thoker2022severe}. For self-supervised pretraining, we use Kinetics-400 or Mini-Kinetics.}
    \label{tab:severe_desc}
    \vspace{-1em}
\end{table*}

\vspace{-0.3em}
\subsection{Tubelet Transformation}
\vspace{-0.3em}
\label{section:transformations}
The tubelet motions are simulated by changing the position of the patch across the frames in a video clip, \textit{i.e.} with translation. 
%Next, we propose to add more variety to the simulated motions with spatial changes in the moving patch. 
%However, in real videos when objects move across time, they may also go through some spatial transformations \eg scale changes when moving in relation to the cameras, inplane rotations, shape changes when performing an action, etc. 
In reality, the motion of objects in space may appear as other transformations in videos, for instance, scale decreasing as the object moves away from the camera or motions due to planer rotations.  Motivated by this, we propose to add more complexity and variety to the simulated motions by transforming the tubelets. In particular, we propose scale, rotation, and shear transformations.
%
%Next, we propose to add more variety to the simulated motions with spatial changes in the moving patch.
As before, we sample keyframes $K$ with the first ($i{=}0$) and last frames ($i{=}T$) always included. Transformations for remaining frames are linearly interpolated. 
Formally, we define a tubelet transformation as a sequence of spatial transformations applied to the patch $p$ in each frame $i$ as:
\begin{align}
\begin{split}
\label{eq:trans_general}
&  \mathrm{Trans}_{F} = [p,F(p,\theta^2),....,..,F(p,\theta^T)], \hspace{2pt}\text{ s.t. } \\ 
& \theta^i = 
\begin{cases}
\mathcal{U}(\mathrm{Min},\mathrm{Max}),  & \text{if } i \in K\\
\mathrm{Interp}(\theta^{k},\theta^{k+1}), & \text{otherwise}
\end{cases}
\end{split}
\end{align}
%\begin{equation}
%\begin{cases}
%\begin{split}
%\label{eq:trans_general}
%\quad 
%&  \mathrm{Trans}_{F} = [p,F(p,f^2),....,..,F(p,f^T)], \hspace{2pt}\text{ s.t. } \\ 
%& f^i \in [Min,Max], \hspace{4pt}\text{if } i \in \mathrm{Keyframes}\\
%& f^i  \in \text{Interpolate}(f_{i}^{k},f_{i}^{k+1}), \hspace{4pt}\text{otherwise}
%\end{split}
%\end{cases}
%\end{equation}
where $F(p,\theta^i)$ applies the transformation to patch $p$ according to parameters $\theta^i$, $\mathcal{U}$ samples from a uniform distribution and  $\theta^k$ and $\theta^{k{+}1}$ are the parameters for the keyframes neighboring frame $i$. % which are sampled randomly from the uniform distribution $\mathcal{U}(\mathrm{Min},\mathrm{Max})$.  %$\mathrm{Min},\mathrm{Max}$ represents the min and max values for the transformation factors.
%As before, multiple keyframes $K$ are selected randomly with the first ($i{=}0$) and  the last frame ($i{=}T$) always included.
For the first keyframe, no transformation is applied thus representing the initial state of the patch $p$. 
%For the other keyframes, $f^i$ is randomly sampled from $[Min, Max]$. 
%The transformation factors for the non-keyframes are linearly interpolated between the neighboring keyframes. 
We instantiate three types of such tubelet transformations: scale, rotation, and shear. Examples are shown in Figure~\ref{fig:transformations}.

\noindent \textbf{Scale.} We scale the patch across time with $F(p, \theta^i)$ and 
%Thus for scale tubelet transformation $F() = Resize()$ and the  transformation  factors $f^i = (w^i,h^i.)$ denote the target scale for the patch in frame $i$ with $ w^i$ being the width and $ h^i$ the height of the patch.
horizontal and vertical scaling factors $\theta^i{=}(w^i,h^i)$. % and $F(p, \theta^i)$ applies the transformation matrix $\big[\begin{smallmatrix} w^i & 0 \\ 0 & h^i \end{smallmatrix}\big]$ to patch $p$. 
%For the first frame, we set the scaling factor to 1, representing the original size of the patch $H^{'},W^{'}$ and use  
To sample $w^i$ and $h^i$, we use $\mathrm{Min}{=}0.5$ and $\mathrm{Max}{=}1.5$. 
%as we randomly select scaling factors between $U(0.5 - 1.5)$ and linearly interpolate the scaling factors for the rest of the frames.  
%Then, Eq.~\eqref{eq:trans_general} can be instantiated with $F(p,f^i) = \mathrm{Resize}(p, (w_i,h^i)$
%\cs{sloppy:}: 
%\begin{equation}
%\textit{Trans}_{scale} = [f(W',H'),..,f(w^m,h^m) ,..,f(w^T,h^T)],
%\mathrm{Trans}_{\mathrm{Resize}} = [P, \mathrm{Resize}(P,s^2),..,\mathrm{Resize}((P,s^T)],
%\end{equation}
%where $s^1 = (W',H')$ is the initial size of the patch, 
%where  $s^i {=} (w^i,h^i.)$ denotes the target scale of the patch in frame $i$ and $ w^i,h^i$ represent the width and height of the patch.
%The left of Figure~\ref{fig:transformations} shows examples of such a transformation. 

\noindent \textbf{Rotation.} In this transformation $F(p, \theta^i)$ applies in-plane rotations to tubelet patches.  
%Thus for rotation tubelet transformation, $F() = Rotate()$  and the  transformation  factors $f^i = r^i$ is the target rotation angle for the patch  in frame $i$. 
Thus, $\theta^i$ is a rotation angle sampled from $\mathrm{Min}{=}{-}90^{\circ}$ and $\mathrm{Max}{=}{+}90^{\circ}$. 
%\begin{equation}
%\mathrm{Trans}_{\mathrm{Rotate}} = [P, \mathrm{Rotate}(P,r^2),..,\mathrm{Rotate}((P,r^T)],
%\end{equation}
%where  $r^i$  is the target rotation angle for the patch  in frame $i$.
%Figure~\ref{fig:transformations} (middle) shows examples of such a rotation transformation.

\noindent \textbf{Shear.} We shear the patch  as the tubelet progresses with $F(p, \theta^i)$.  
%Then, $F() = Shear()$ is the transformation function with  factors $f^i {=} (s^{i}_{x},s^{i}_{y}.)$ denoting the target shearing factor for the patch  in frame $i$. $s^{i}_{x},s^{i}_{y}$ represents the shear along the X and Y axes respectively. 
The shearing parameters are $\theta^i{=}(r^i,s^{i})$ which are sampled using $\mathrm{Min}{=}{-}1.5$ and $\mathrm{Max}{=}1.5$. 
%Again, we set the shear factor to 0 for the first frame representing the original shape of the patch, and  sample as per Eq.~\eqref{eq:trans_general} for other frames.
%\begin{equation}
%\mathrm{Trans}_{\mathrm{Shear}} = [P, \mathrm{Shear}(P,s^2),..,\mathrm{Shear}((P,s^T)],
%\end{equation}
%where  $s^i {=} (s^{i}_{x},s^{i}_{y}.)$ denotes the target shearing factor for the patch  in frame $i$ and $s^{i}_{x},s^{i}_{y}$ represents the shear along the X and Y axes respectively. 
%Figure~\ref{fig:transformations} (right column) shows examples of the shear transformation.

With these tubelet transformations and the motions created in Section~\ref{section:motions} we are able to simulate a variety of subtle motions in videos, making the model data-efficient. By learning the similarity between the same tubelet overlaid onto different videos, our model pays less attention to spatial features, instead learning to represent these subtle motions. This makes the learned representation generalizable to different domains and action granularities.  

%\cs{This paragraph feels repetitive, can probably go. Instead we need a short closing statement for the section.}
%To summarize, for tubelet-contrastive learning  we first sample two video clips from two different videos and crop a set of patches from one of the clips. From the patches, we construct a set of tubelets with linear or non-linear motions. Each tubelet is then transformed by one of the transformation functions.  The resulting set of  tubelets is overlaid ($\odot $) on the original clips to create positive pairs that have the same motion patterns. By contrasting such positive pairs, the network is optimized to learn similarities between clips with the same motion-tublets and dissimilarities between clips with different motion tublets. Since the motion-tubelets are the only common information between the positive pairs, such optimization forces the network to rely on temporal cues when solving this contrastive task, which in turn results in learning motion-focused feature representations for videos. 

%% file: sections/4_expirements.tex
\vspace{-0.3em}
\section{Experiments}
\vspace{-0.3em}
\subsection{Datasets, Evaluation \& Implementation}
\vspace{-0.3em}

\noindent\textbf{Pretraining Datasets. } 
Following prior work \cite{videomoco-pan2021videomoco,gdt-patrick2020multimodal,dave2022tclr,ctp-wang2021unsupervised,rspnet-chen2020RSPNet,huang2021ascnet}  we use \textbf{Kinetics-400}~\cite{Kinetics-400-arxiv} for self-supervised pretraining. Kinetics-400 is a large-scale action recognition dataset containing 250K videos of 400 action classes.
To show data efficiency, we also pretrain with \textbf{Mini-Kinetics}~\cite{mini_kinetics}, a subset containing 85K videos of 200 action classes. %For both cases, we use the training set without action labels for self-supervised pretraining. 

\noindent \textbf{Downstream Evaluation. } 
To evaluate the video representations learned by our tubelet contrast, we finetune and evaluate our model on various downstream datasets summarized in Table~\ref{tab:severe_desc}.
Following previous self-supervised work, we evaluate on standard benchmarks: \textbf{UCF101}~\cite{UCF-101-arxiv} and \textbf{HMDB51}~\cite{HMDB-51-ICCV}. These action recognition datasets contain coarse-grained actions with domains similar to Kinetics-400. 
%UCF101 contains 13,320 videos from 101 action classes while HMDB51 contains 6,890 videos from 51 action classes. 
For both, we report top-1 accuracy on split 1 from the original papers.
We examine the generalizability of our model with the \textbf{SEVERE} benchmark proposed in our previous work~\cite{thoker2022severe}. This consists of eight experiments over four downstream generalization factors: \textit{domain shift}, \textit{sample efficiency}, \textit{action granularity}, and \textit{task shift}. %SEVERE contains two experimental setups for each factor.
\textit{Domain shift} is evaluated on Something-Something v2~\cite{SS-v2-arxiv} (SSv2) and FineGym~\cite{Gym-99-arxiv} (Gym99)  
which vary in domain relative to Kinetics-400. 
\textit{Sample efficiency} evaluates low-shot action recognition on UCF101~\cite{UCF-101-arxiv} and FineGym~\cite{Gym-99-arxiv} with 1,000 training samples, referred to as UCF ($10^{3}$) and Gym ($10^{3}$).
\textit{Action granularity} evaluates semantically similar actions using FX-S1 and UB-S1 subsets from  FineGym~\cite{Gym-99-arxiv}. In both subsets, action classes belong to the same element of a gymnastic routine, \eg, FX-S1 is types of jump.  
\textit{Task shift} evaluates tasks beyond single-label action recognition. Specifically, it uses temporal repetition counting on UCFRep~\cite{ucfrep-zhang2020context}, a subset of UCF-101~\cite{ucfrep-zhang2020context}, and multi-label action recognition on Charades~\cite{charades-sigurdsson:hal-01418216}. %where videos are an average of 30 seconds long. 
The experimental setups are detailed in Table~\ref{tab:severe_desc} and all follow SEVERE~\cite{thoker2022severe}. 

\noindent\textbf{Tubelet Generation and Transformation.} Our clips are 16 $112 {\times} 112$ frames with standard spatial augmentations: random crops, horizontal flip, and color jitter. %Accordingly, we set $T{=}16$, $W{=}112$ and $H{=}112$ for tubelet generation.  
We randomly crop 2 patches to generate $M{=}2$ tubelets (Eq.~\ref{eq:overlay_M_tubelets}). The patch size $H^{'}{\times} W^{'}$ is uniformly sampled from $[16{\times}16, 64{\times64}]$. We also randomly sample a patch shape from a set of predefined shapes. For linear motions, we use $\Delta{=}[40{-}80]$ displacement difference. For non-linear motion, we use $N{=}48$ and a smoothing factor of $\sigma{=}8$ (Eq.~\ref{eq:nonlinear_motion}). For linear motion and all tubelet transformations, we use $K{=}3$ keyframes.

\noindent\textbf{Networks, Pretraining and Finetuning.}
We use R(2+1)D-18~\cite{tran2018closer} as the video encoder, following previous self-supervision works~\cite{pace,gdt-patrick2020multimodal,rspnet-chen2020RSPNet,fame,dave2022tclr,videomoco-pan2021videomoco}. The projection head is a 2-layer MLP with 128D output. We use momentum contrast~\cite{moco} to increase the number of negatives $|\mathcal{N}|$ (Eq.~\ref{eq:tcl}) to 16,384 for Mini-Kinetics and 65,536 for Kinetics. 
We use temperature $\tau{=}0.2$ (Eq.~\ref{eq:tcl}). The model is optimized using SGD with momentum 0.9,  learning rate 0.01, and weight decay 0.0001. We use a batch size of 32 for Mini-Kinetics and 128 for Kinetics, a cosine scheduler~\cite{cosine_sch}, and pretrain for 100 epochs. After pretraining, we replace the projection head with a task-dependent head as in SEVERE~\cite{thoker2022severe} and finetune the whole network with labels for the downstream task.  We provide finetuning details in the supplementary. %

%% file: sections/5_expirements.tex
\vspace{-0.5em}
\subsection{Ablation Studies \& Analysis}
\vspace{-0.3em}
To ablate the effectiveness of individual components we pretrain on Mini-Kinetics and evaluate on UCF ($10^{3}$), Gym ($10^{3}$), Something-Something v2 and UB-S1. To decrease the finetuning time we use a subset of Something Something (SSv2-Sub) with 25\% of the training data (details in supplementary). Unless specified otherwise, we use non-linear motion and rotation to generate tubelets.
%\cs{what video encoder is used?}

\begin{table}
    \centering
    \setlength{\tabcolsep}{2pt}
    \resizebox{\linewidth}{!}{
    \begin{tabular}{lcccc} \toprule  
    %Pre-training& UCF ($10^{3}$) & Gym ($10^{3}$) &UB-S1 & SSv2-Sub\\
    & UCF ($10^{3}$) & Gym ($10^{3}$) & SSv2-Sub & UB-S1 \\
         \midrule
         \rowcolor{Gray}
         \textbf{Temporal Contrast} & & & &\\
         Baseline    & 57.5 & 29.5        & 44.2  & 84.8      \\
         \rowcolor{Gray}
         \textbf{Tubelet Contrast} & & & &\\
         Tubelet Generation & 48.2  &28.2    & 40.1  & 84.1      \\
         Tubelet Motion
          & 63.0 & 45.6          & 47.5  & 90.3      \\
         Tubelet Transformation  & 65.5 & 48.0 & 47.9  & 90.9      \\
        \bottomrule
    \end{tabular}}
    \vspace{-0.8em}
    \caption{\textbf{Tubelet-Contrastive Learning} considerably outperforms temporal contrast on multiple downstream settings. Tubelet motion and transformations are key. % to this improvement. %Each component contributes to the improvement of downstream performance.
    }
    \label{tab:ablation_main}
    \vspace{-1em}
\end{table}

\noindent\textbf{Tubelet-Contrastive Learning.}
Table~\ref{tab:ablation_main} shows the benefits brought by our tubelet-contrastive learning. We first observe that our full tubelet-contrastive model improves considerably over the temporal contrastive baseline, which uses MoCo~\cite{moco} with a temporal crop augmentation. This improvement applies to all downstream datasets but is especially observable with Gym ($10^{3}$) (+18.5\%) and UB-S1 (+6.1\%) where temporal cues are crucial. Our model is also effective on UCF ($10^{3}$) (+8.0\%) where spatial cues are often as important as temporal ones. These results demonstrate that learning similarities between synthetic tubelets produces generalizable, but motion-focused, video representations required for finer temporal understanding.

It is clear that the motion within tubelets is critical to our model's success as contrasting static tubelets obtained from our tubelet generation (Section~\ref{section:tbcl}) actually decreases the performance from the temporal contrast baseline. 
%\cs{Long sentence, hard to parse:}
%This is because positive pairs with such tubelets only share spatial similarity in the form of a static patch that is duplicated in all frames of the video clips at the same location, without any motion and therefore does not require any temporal understanding for learning similarities which results in  poor video feature representations. 
%
%This demonstrates that tubelet motion is key to learning effective video representations. 
When tubelet motion is added (Section~\ref{section:motions}), performance improves considerably, \eg, Gym ($10^{3}$) +17.4\% and SSv2-Sub +7.4\%. 
Finally, adding more motion types via tubelet transformations (Section~\ref{section:transformations}) further improves the video representation quality, \eg, UCF ($10^{3}$) +2.5\% and  Gym ($10^{3}$) +2.4\%. This highlights the importance of including a variety of motions beyond what is present in the pretraining data to learn generalizable video representations. %Further validating our idea that adding such motion patterns for learning improves downstream performance. 

\begin{table}
    \centering
    \resizebox{0.95\linewidth}{!}{
    \begin{tabular}{lcccc} \toprule Tubelet Motion & UCF ($10^{3}$) & Gym ($10^{3}$) &  SSv2-Sub & UB-S1  \\
         \midrule
         No motion & 48.2  &28.2      & 40.1 & 84.1 \\
         Linear   & 55.5 & 34.6       & 45.3 & 88.5 \\
         Non-Linear   & 63.0 & 45.6   & 47.5 & 90.3 \\
        \bottomrule
    \end{tabular}}
    \vspace{-0.8em}
    \caption{\textbf{Tubelet Motions.} Learning from tubelets with non-linear motion benefits multiple downstream settings.}
    \label{tab:ablation_motions}
    \vspace{-0.8em}
\end{table}

 \begin{table}
    \centering
    \resizebox{0.95\linewidth}{!}{
    \begin{tabular}{lcccc} \toprule Transformation & UCF ($10^{3}$) & Gym ($10^{3}$)  & SSv2-Sub &UB-S1\\
         \midrule
         None   & 63.0 & 45.6 & 47.5 & 90.5    \\
         Scale  & 65.1 & 46.5 & 47.0 & 90.5   \\
         Shear  & 65.2 & 47.5 & 47.3 & 90.9   \\
         Rotation & 65.5 & 48.0 & 47.9 & 90.9   \\
        \bottomrule
    \end{tabular}}
    \vspace{-0.8em}
    \caption{\textbf{Tubelet Transformation.} Adding motion patterns to tubelet-contrastive learning through transformations improves downstream performance. Best results for rotation.}
    \label{tab:ablation_transformations}
    \vspace{-0.8em}
\end{table}

\noindent\textbf{Tubelet Motions.} Next, we ablate the impact of the  tubelet motion type (Section~\ref{section:motions}) without transformations. We compare the performance of static tubelets with no motion, linear motion, and non-linear motion in Table~\ref{tab:ablation_motions}. Tubelets with simple linear motion already improve performance for all four datasets, \eg, +6.4\% on Gym ($10^3$). %\cs{highlight one result?}. 
Using non-linear motion further improves results, for instance with an additional +11.0\% improvement on Gym ($10^3$). %\cs{highlight another result?}. 
We conclude that learning from non-linear motions provides more generalizable video representations. 
%This demonstrates the need to simulate a variety of motion patterns to learn video representations that can apply to a variety of downstream domains. %As a result, we use tubelets with non-linear motions in our final model.
 %\cs{It would be interesting to look at individual action recognition results to see whether there are any spectacular individual results, both pro and con. These could be reported in supplemental if of interest.}

\begin{table}
    \centering
    \resizebox{0.88\linewidth}{!}{
    \begin{tabular}{lcccc}
         \toprule \#Tubelets & UCF ($10^{3}$) & Gym ($10^{3}$)  & SSv2-Sub & UB-S1\\
         \midrule
         1 & 62.0 & 39.5 &  47.1 & 89.5 \\
         2 & 65.5 & 48.0 &  47.9 & 90.9 \\
         3 & 66.5 & 46.0 &  47.5 & 90.9 \\
        \bottomrule
    \end{tabular}}
    \vspace{-0.8em}
    \caption{\textbf{Number of Tubelets.} Overlaying two tubelets in positive pairs improves downstream performance. %Best results for two tubelets. %\cs{We can keep the ablation, but move the table with detailed numbers to supplemental?}
    }
    \label{tab:ablation_num_tubelets}
    \vspace{-1em}
\end{table}

\begin{figure}
\centering
\includegraphics[width=\linewidth]{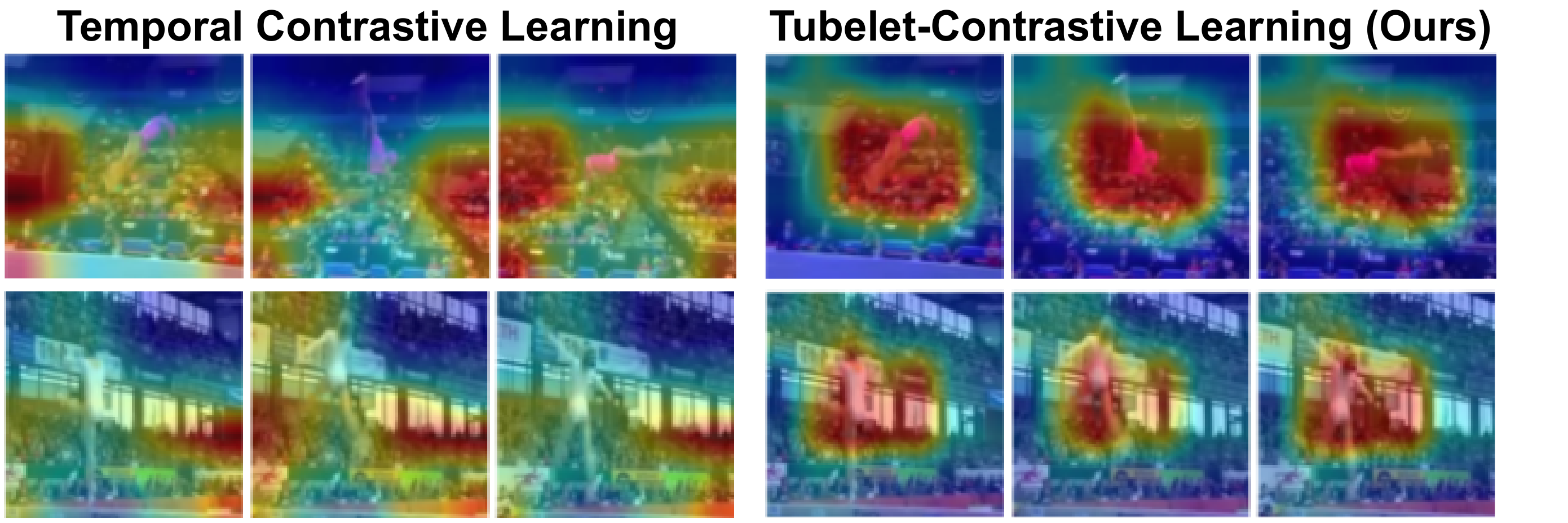}
\vspace{-2em}
\caption{\textbf{Class-Agnostic Activation Maps without Finetuning} for the temporal contrastive baseline and our tubelet-contrast.  Our model better attends to regions with motion. }
\vspace{-0.8em}
\label{fig:caam}
\end{figure}

\begin{table}[t]
    \centering
    %\midsepremove
    \setlength{\tabcolsep}{5pt}
    \resizebox{0.88\linewidth}{!}{\begin{tabular}
     {l  cccc}
    \toprule
        & \multicolumn{2}{Sc}{\textbf{Linear Classification}}   & \multicolumn{2}{Sc}{\textbf{ Finetuning }}  \\ 
      \cmidrule(lr){2-3} \cmidrule(lr){4-5} 
      %\addlinespace[0.1cm]
         &   \multicolumn{1}{c}{UCF101} & \multicolumn{1}{c}{Gym99}  & \multicolumn{1}{c}{UCF101} & \multicolumn{1}{c}{Gym99}  \\ 
         \midrule
         Temporal Contrast &  \textbf{58.9}   & 19.7  &87.1   & 90.8  \\ 
         Tubelet Contrast  & 30.0   & \textbf{34.1} &  \textbf{91.0}   & \textbf{92.8} \\ 
        %\addlinespace[0.01cm]
         \bottomrule
    \end{tabular}
    }
    \vspace{-0.8em}
    \caption{\textbf{Appearance vs Motion}. Our method learns to capture motion dynamics with pretraining and can easily learn appearance features with finetuning.}
\label{tab:qa}
\vspace{-1em}
\end{table}

\begin{figure*}[t!]
\centering
\includegraphics[width=0.98\linewidth]{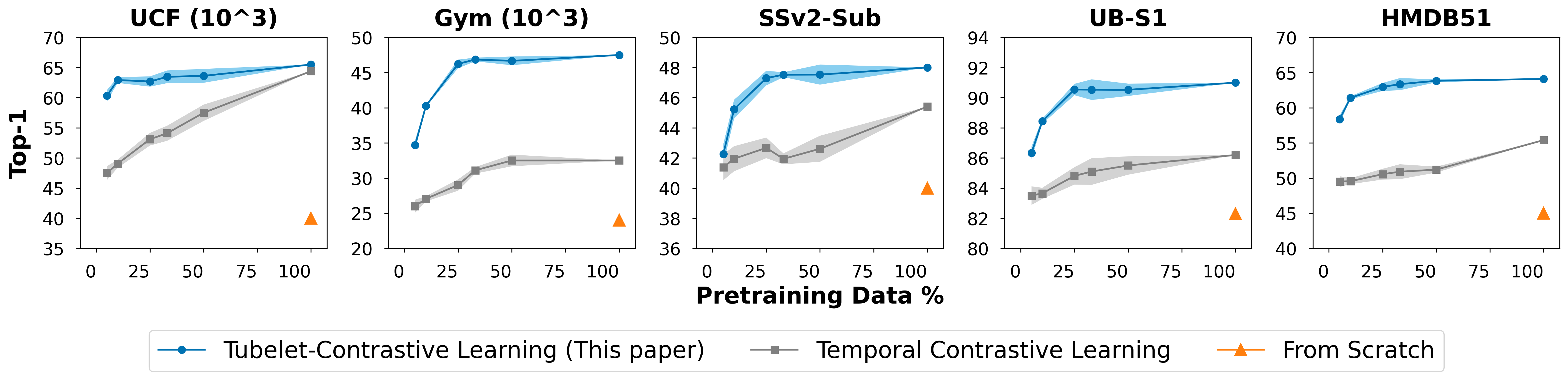}
\vspace{-1.2em}
\caption{\textbf{Video-Data Efficiency of Tubelet-Contrastive Learning.} Our approach maintains performance when using only 25\% of the pretraining data. When using 5\% of the pretraining data, our approach is still more effective than using 100\% with the baseline for Gym ($10^3$), UB-S1, and HMDB51. Results are averaged over three pretraining runs with different seeds.%\hd{Typo in figure legend, temoral $\rightarrow$ temporal} 
%\hd{Replace FX-S1 with UB-S1}
\vspace{-1em}
}
\label{fig:data_efficiency}
\end{figure*}

\noindent\textbf{Tubelet Transformation.}
Table~\ref{tab:ablation_transformations} compares the  proposed tubelet transformations (Section~\ref{section:transformations}). % when using non-linear motion.
%\cs{Do not write this from the dataset perspective but from the transformation perspective:}
All four datasets benefit from transformations, with rotation being the most effective. %We find that scaling improves results on UCF ($10^3$) (+2.1\%) and Gym ($10^3$) (+0.9\%) while maintaining performance on UB-S1 and SSv2-Sub. Shearing gives better performance over scaling on all datasets, \eg, Gym ($10^3$) (+1.0\%) however, performance for SSv2-Sub is still similar to using no transformation. Rotation is the most effective transformation, increasing performance on all four datasets.
%For UCF($10^{3}$) and Gym ($10^{3}$) all three transformations are helpful. For UB-S1, both shearing and rotation improve performance, while for SSv2-Sub only rotations are helpful. 
The differences in improvement for each transformation are likely due to the types of motion present in the downstream datasets. For instance, Gym ($10^3$) and UB-S1 contain gymnastic videos where actors are often spinning and turning but do not change in scale due to the fixed camera, therefore rotation is more helpful than scaling. %While SomethingSomething contains videos recorded with static cameras that have motion types like moving objects from left to right or top to bottom, etc. %This is probably because scale mostly adds spatial variations to the tubelets while rotation and shearing add both spatial and  motion variations to  the tubelets resulting in learning better spatial-temporal dynamics. 
We also experiment with combinations of transformations in supplementary but observe no further improvement. %Thus,  we choose rotation as the tubelet transformation for our final model.   

%\vspace{-0.20em}
\noindent\textbf{Number of Tubelets.} %\cs{Reduce amount of text and move all details incl table to appendix to win space?}
%Finally, we investigate the number of tubelets overlaid on video clips to generate positive pairs (Eq.~\ref{eq:overlay_M_tubelets}). Table~\ref{tab:ablation_num_tubelets} shows the comparison for using one, two, and three tubelets per video. 
We investigate the number of tubelets used in each video in Table~\ref{tab:ablation_num_tubelets}. One tubelet is already more effective than temporal contrastive learning, \eg, 29.5\% vs. 39.5\% for Gym ($10^3$). Adding two tubelets improves accuracy on all datasets, \eg, +8.5\% for Gym ($10^3$). %This improvement starts to saturate when using three tubelets, with only UCF ($10^3$) benefitting. Therefore, we use two tubelets in our final model.

\noindent\textbf{Analysis of Motion-Focus. } 
To further understand what our model learns, Figure~\ref{fig:caam} visualizes the class agnostic activation maps~\cite{CAAM} without finetuning for the baseline and our approach. We observe that even  without previously seeing any FineGym data, our approach attends better to the motions than the temporal contrastive baseline, which attends to the background regions. This observation is supported by the linear classification and finetuning results on  
UCF101 (appearance-focused) and Gym99 (motion-focused) in Table~\ref{tab:qa}. When directly  predicting from the learned features with linear classification, our model is less effective than temporal contrast for appearance-based actions in UCF101, but positively affects actions requiring fine-grained motion understanding in Gym99. With finetuning, our tubelet-contrastive representation is able to add spatial appearance understanding and maintain its ability to capture temporal motion dynamics, thus it benefits both UCF101 and Gym99.

\vspace{-0.8em}
\subsection{Video-Data Efficiency} 
\vspace{-0.6em}
To demonstrate our method's data efficiency, we pretrain using subsets of the Kinetics-400. In particular, we sample $5\%,10\%,25\%,33\% \text{ and } 50\%$ of the Kinetics-400 training set with three random seeds and pretrain our model and the temporal contrastive baseline. %We repeat the experiment three times to sample different subsets of data with different seeds. 
We compare the effectiveness of these representations after finetuning on UCF ($10^{3}$), Gym($10^{3}$), SSv2-Sub, UB-S1, and HMDB51 in Figure~\ref{fig:data_efficiency}.
On all downstream setups, our method maintains similar performance when reducing the pretraining data to just 25\%, while the temporal contrastive baseline performance decreases significantly. Our method is less effective when using only 5\% or 10\% of the data, but remarkably still outperforms the baseline trained with 100\% data for Gym ($10^3$), UB-S1, and HMDB. We attribute our model's data efficiency to the tubelets we add to the pretraining data. In particular, our non-linear motion and transformations generate a variety of synthetic tubelets that 
simulate a greater variety of fine-grained motions than are present in the original data.

%% file: sections/6_expirements.tex
\vspace{-0.6em}
\subsection{Standard Evaluation: UCF101 and HMDB51}
\vspace{-0.5em}
We first show the effectiveness of our proposed method on standard coarse-grained action recognition benchmarks UCF101 and HMBD51, where we compare with prior video self-supervised works. For a fair comparison, we only report methods in Table~\ref{tab:ucf_hmdb_sota} that use the R(2+1)D-18 backbone and Kinetics-400 as the pretraining dataset. 

%\noindent\textbf{RGB-only Comparison.} 
First, we observe that %among the self-supervised methods that only use the RGB modality for pretraining, 
our method obtains the best results for UCF101 and HMDB51. % with R(2+1)D-18 backbone. 
 The supplementary material shows we also achieve similar improvement with the R3D and I3D backbones. In particular, with R(2+1)D our method beats CtP~\cite{ctp-wang2021unsupervised} by 2.6\% and 2.4\%, TCLR~\cite{dave2022tclr} by 2.8\% and 4.1\%, and TE~\cite{jenni2021time_eqv} by 2.8\% and 1.9\% all of which aim to learn finer temporal representations.   This confirms that explicitly contrasting tubelet-based motion patterns   
 results in a better video representation than learning temporal distinctiveness or prediction.  We also outperform FAME~\cite{fame} by 6.2\% and 9.6\% on UCF101 and HMDB51. FAME aims to learn a motion-focus representation by pasting the foreground region of one video onto the background of another to construct positive pairs for contrastive learning. We however are not limited by the motions present in the set of pretraining videos as we simulate new motion patterns for learning. We also outperform prior multi-modal works which incorporate audio or explicitly learn motion from optical flow. 
Since our model is data-efficient, we can pretrain on Mini-Kinetics and still outperform all baselines which are trained on the 3x larger Kinetics-400. 

 \begin{table}[t!]
\centering
    \setlength{\tabcolsep}{6pt}
\resizebox{0.92\linewidth}{!}{
    \begin{tabular}{llcc}
        \toprule
        \textbf{Method} & \textbf{Modality} & UCF101 & HMDB51\\
         %& & UCF101 & HMDB51 \\\hline
%         \midrule
        
        \midrule
        %Pace Prediction~\cite{pace} & RGB &   77.1 & 36.6 \\
        %MoCo~\cite{moco} & R(2+1)D & Kinetics-400 & 32 & V &  78.7 & 49.2 \\
        VideoMoCo~\cite{videomoco-pan2021videomoco} & RGB &   78.7 & 49.2 \\
        RSPNet~\cite{rspnet-chen2020RSPNet} & RGB  &  81.1 &44.6 \\
        SRTC~\cite{srtc} & RGB &    82.0 & 51.2 \\
        FAME~\cite{fame}  & RGB &    {84.8}	& {53.5} \\
        MCN~\cite{mcn} & RGB     & 84.8 & 54.5 \\
        \color{gray}
        AVID-CMA~\cite{avid-cma-morgado2021audio} & \color{gray} RGB+Audio & \color{gray} 87.5 & \color{gray} 60.8 \\
        TCLR~\cite{dave2022tclr} & RGB &   88.2 &60.0 \\
        TE~\cite{jenni2021time_eqv} & RGB & 88.2 &	62.2\\	
        CtP~\cite{ctp-wang2021unsupervised} & RGB & {88.4}	& {61.7} \\
        \color{gray}
        MotionFit~\cite{motion_fit} & \color{gray} RGB+Flow &   \color{gray} 88.9 & \color{gray} 61.4 \\ 
        \color{gray} GDT~\cite{gdt-patrick2020multimodal} & \color{gray} RGB+Audio &  \color{gray} 89.3 & \color{gray} 60.0 \\
        
        \midrule
        %\rowcolor{visualblue}
        \textbf{\textit{This paper $^\dagger$}} & RGB &  \textbf{90.7}	& \textbf{65.0} \\
        %\rowcolor{visualblue}
        \textbf{\textit{This paper}} & RGB & \textbf{91.0}	& \textbf{64.1} \\
        \bottomrule
    \end{tabular}
    }
    \vspace{-0.9em}
    \caption{\textbf{Standard Evaluation: UCF101 and HMDB51} using R(2+1)D. 
    %All models utilize Kinetics-400 for self-supervised pretraining. 
    Gray lines indicate use of additional modalities during self-supervised pretraining. Note that our method pretrained on Mini-Kinetics ($\dagger$) outperforms all methods which pretrain on the 3${\times}$ larger Kinetics-400.
    }
    \vspace{-1em}
    \label{tab:ucf_hmdb_sota}
\end{table}

\begin{table*}[t!]
    \centering
    \midsepremove
    \setlength{\tabcolsep}{5pt}
    \resizebox{\linewidth}{!}{\begin{tabular}
     {l c ccc cc ccccc}
    \toprule
    
      \addlinespace[0.1cm]
        & & \multicolumn{2}{Sc}{\textbf{Domains}}   & \multicolumn{2}{Sc}{\textbf{Samples}}  & \multicolumn{2}{Sc}{\textbf{Actions}} & \multicolumn{2}{Sc}{\textbf{Tasks}}\\
      \cmidrule(lr){3-4} \cmidrule(lr){5-6} \cmidrule(lr){7-8} \cmidrule(lr){9-10}
         %\multicolumn{1}{l}{\textbf{Methods}}  \\ 
      \addlinespace[0.1cm]
         &  \multicolumn{1}{c}{Backbone} & \multicolumn{1}{c}{SSv2} & \multicolumn{1}{c}{Gym99}  & \multicolumn{1}{c}{UCF ($10^{3}$)} & \multicolumn{1}{c}{Gym ($10^{3}$)}  & \multicolumn{1}{c}{FX-S1 } & \multicolumn{1}{c}{UB-S1}&  \multicolumn{1}{c}{UCF-RC$\downarrow$} &  \multicolumn{1}{c}{Charades} & \textbf{Mean } & \textbf{Rank$\downarrow$}\\
         \midrule
           %\addlinespace[0.01cm]
        \color{gray} SVT~\cite{svt} & \color{gray}ViT-B  & \color{gray}59.2   & \color{gray}62.3   & \color{gray}83.9   & \color{gray}18.5   & \color{gray}35.4   & \color{gray}55.1 & \color{gray}0.421 & \color{gray}35.5   &  \color{gray}51.0                     & \color{gray}~8.9            \\
        \color{gray} VideoMAE~\cite{tong2022videomae}  &\color{gray}ViT-B           & \color{gray}69.7   & \color{gray}85.1   & \color{gray}77.2   & \color{gray}27.5   & \color{gray}37.0   & \color{gray}78.5 & \color{gray}0.172 & \color{gray}12.6   &                     \color{gray}58.1 & \color{gray}~8.3            \\
 %\color{gray}        MME~\cite{sun2023mme}     &\color{gray}ViT-B          & \color{gray}70.5   & \color{gray}91.7   & \color{gray}84.4   & \color{gray}41.4   & \color{gray}69.7   & \color{gray}90.1   & \color{gray}0.174 & \color{gray}23.6   &  \color{gray}69.3                    & -           \\
         \midrule
         Supervised~\cite{tran2018closer}  &R(2+1)D-18             & 60.8   & 92.1   & 86.6   & 51.3   & 79.0   & 87.1 & 0.132 & 23.5   &  70.9                     & 3.9            \\
         \midrule
         None   &R(2+1)D-18                  & 57.1   & 89.8   & 38.3   & 22.7   & 46.6   & 82.3 & 0.217 & ~7.9    &  52.9                         &11.6            \\
         SeLaVi~\cite{selavi}   &R(2+1)D-18               & 56.2   & 88.9   & 69.0   & 30.2   & 51.3   & 80.9 & 0.162 & ~8.4    & 58.6                      & 11.0           \\
         MoCo~\cite{moco}     &R(2+1)D-18                & 57.1   & 90.7   & 60.4   & 30.9   & 65.0   & 84.5 & 0.208 & ~8.3    & 59.5                       & ~9.1            \\
         VideoMoCo~\cite{videomoco-pan2021videomoco}  &R(2+1)D-18              & 59.0   & 90.3   & 65.4   & 20.6   & 57.3   & 83.9 & 0.185 & 10.5   & 58.6                       & ~9.1            \\
         Pre-Contrast~\cite{pretext-contrast}   &R(2+1)D-18      & 56.9   & 90.5   & 64.6   & 27.5   & 66.1   & 86.1 & 0.164 & ~8.9    & 60.5                      & ~9.0            \\
         AVID-CMA~\cite{avid-cma-morgado2021audio}  &R(2+1)D-18               & 52.0   & 90.4   & 68.2   & 33.4   & 68.0   & 87.3 & 0.148 & ~8.2    & 61.6                       & ~9.0            \\
         GDT~\cite{gdt-patrick2020multimodal}  &R(2+1)D-18                    & 58.0   & 90.5   & \textbf{78.4}   & 45.6   & 66.0   & 83.4 & \textbf{0.123} & ~8.5    & 64.8    & ~8.6           \\
         RSPNet~\cite{rspnet-chen2020RSPNet}    &R(2+1)D-18               & 59.0   & 91.1   & 74.7   & 32.2   & 65.4   & 83.6 & 0.145 & ~9.0    & 62.6                       & ~8.0           \\
                  TCLR~\cite{dave2022tclr}    &R(2+1)D-18                 & 59.8   & 91.6   & 72.6   & 26.3   & 60.7   & 84.7 & 0.142 & \textbf{12.2}   & 61.7              & ~7.6            \\
         CtP~\cite{ctp-wang2021unsupervised}   &R(2+1)D-18                   & 59.6   & 92.0   & 61.0   & 32.9   & 79.1   & 88.8 & 0.178 & ~9.6    & 63.2                     & ~5.6            \\
         \midrule                                                                  \addlinespace[0.03cm]                          %                  \rowcolor{visualblue}       
         %\rowcolor{audiogreen}
         \textbf{\textit{This paper $^\dagger$}}    &R(2+1)D-18                  & {59.4}   & {92.2}   & {65.5}   & \textbf{48.0}   & {78.3}   & 90.9 & {0.150}  & {9.0}   & 66.0  &~5.4    \\
                  \textbf{\textit{This paper}}  &R(2+1)D-18                    & \textbf{60.2}   & \textbf{92.8}   & {65.7}   & 47.0   & \textbf{80.1}   & \textbf{91.0} & {0.150}  & {10.3}   & \textbf{66.5}  & ~\textbf{4.1}    \\
        \addlinespace[0.01cm]
         \bottomrule
    \end{tabular}
    }
    \vspace{-0.9em}
         \caption{\textbf{SEVERE Generalization Benchmark.}
         Comparison with prior self-supervised methods for generalization to downstream domains, fewer samples, action granularity, and tasks. $\downarrow$ indicates lower is better. Results for baselines are taken from SEVERE \cite{thoker2022severe}.
         Our method generalizes best, even when using the 3x smaller Mini-Kinetics dataset ($\dagger$) for pretraining.}
    \label{tab:severe-performance}
    \vspace{-1.0em}
\end{table*}

\vspace{-0.5em}
\subsection{SEVERE Generalization Benchmark} 
\vspace{-0.5em}
Next, we compare to prior works on the challenging SEVERE benchmark \cite{thoker2022severe}, which evaluates video representations for generalizability in \textit{domain shift}, \textit{sample efficiency}, \textit{action granularity}, and \textit{task shift}. We follow the same setup as in the original SEVERE benchmark and use an R(2+1)D-18 backbone pretrained on Kinetics-400 with our tubelet-contrast before finetuning on the different downstream settings.
Results are shown in Table~\ref{tab:severe-performance}.

\noindent \textbf{Domain Shift.} Among the evaluated methods our proposal achieves the best results on SSv2 and Gym99. These datasets differ considerably from Kinetics-400, particularly in regard to the actions,  environment and viewpoint. Our improvement demonstrates that the representation learned by our tubelet-contrast is robust to various domain shifts. 

\vspace{-0.05em}
\noindent \textbf{Sample Efficiency.}
For sample efficiency, we achieve a good gain over all prior works on Gym ($10^{3}$), \eg, +20.7\% over TCLR~\cite{dave2022tclr} and +14.1\%  over CtP~\cite{ctp-wang2021unsupervised}. Notably, the gap between the second best method GDT~\cite{gdt-patrick2020multimodal} and all others is large, demonstrating the challenge. For UCF ($10^{3}$), our method is on par with VideoMoCo\cite{videomoco-pan2021videomoco} and CtP but is outperformed by GDT and RSPNet~\cite{rspnet-chen2020RSPNet}. %\etc. 
This is likely due to most actions in UCF101 requiring more spatial than temporal understanding, thus it benefits from the augmentations used by GDT and RSPNet. Our motion-focused representation  requires more finetuning samples on such datasets. 

\vspace{-0.05em}
\noindent \textbf{Action Granularity.} For fine-grained actions in FX-S1 and UB-S1, our method achieves the best performance, even outperforming supervised Kinetics-400 pretraining.
 We achieve a considerable improvement over other RGB-only models, \eg, +19.6\% and +6.3\% over TCLR, as well as audio-visual models, \eg, +14.1\% and +7.6\% over GDT. 
These results demonstrate that the video representation learned by our method are better suited  to fine-grained actions than existing self-supervised methods. We additionally report results on  Diving48~\cite{diving48} in the supplementary.

\vspace{-0.05em}
\noindent \textbf{Task Shift.} 
For the task shift to repetition counting, our method is on par with AVID-CMA~\cite{avid-cma-morgado2021audio} and RSPNet, but worse than GDT. For multi-label action recognition on Charades, our approach is 3rd, comparable to VideoMoCo but worse than TCLR.
 This suggests the representations learned by our method are somewhat transferable to tasks beyond single-label action recognition. However, the remaining gap between supervised and self-supervised highlights the need for future work to explore task generalizability further.
 
\vspace{-0.05em}
\noindent \textbf{Comparison with Transformers.} Table~\ref{tab:severe-performance} also contains recent transformer-based self-supervised works SVT~\cite{svt} and VideoMAE~\cite{tong2022videomae}. We observe that both SVT and VideoMAE have good performance with large amounts of finetuning data (SSv2), in-domain fine-tuning (UCF($10^3$)), and multi-label action recognition (Charades). However, they considerably lag in performance for motion-focused setups Gym99, FX-S1, UB-S1, and repetition counting compared to our tubelet contrast with a small CNN backbone. 
 %Our method also often outperforms MME~\cite{sun2023mme} which extracts  motion information from optical flow and HOG features for pretraining.

\vspace{-0.05em}
\noindent \textbf{Overall SEVERE Performance.} 
Finally, we compare the mean and the average rank across all generalizability factors. Our method has the best mean performance (66.5) and achieves the best average rank (4.1). When pretraining with the 3x smaller Mini-Kinetics our approach still achieves impressive results. We conclude our method improves the generalizability of video self-supervised representations across these four downstream factors while being data-efficient.

%% file: sections/7_conclusion.tex
\vspace{-0.5em}
\section{Conclusion}
\vspace{-0.5em}
This paper presents a contrastive learning method to learn motion-focused video representations in a self-supervised manner. Our model adds synthetic tubelets to videos so that the only similarities between positive pairs are the spatiotemporal dynamics of the tubelets. By altering the motions of these tubelets and applying transformations we can simulate motions not present in the pretraining data. Experiments show that our proposed method is data-efficient and more generalizable to new domains and fine-grained actions than prior self-supervised methods. 

\vspace{0.25em}
\noindent\textbf{Acknowledgements}. This work is part of the research programme Perspectief EDL with project number P16-25 project 3, which is financed by the Dutch Research Council (NWO) domain Applied and Engineering/Sciences (TTW). We thank Piyush Bagad for help with experiments and Artem Moskalev for useful discussions.

%% file: sections/appendix.tex
\clearpage
\noindent\textbf{\Large{Appendix}}
%\begin{center}
%\textbf{Tubelet-Contrastive Self-Supervision for Video-Efficient Generalization\\Supplementary Material}
%\end{center}
% \begin{equation}
%\mathcal{L}(V1,V2) = -\log \frac{\displaystyle \mathrm{exp}(Z_{v_{1}^{'}}\cdot Z_{v_{2}^{'}}/\tau)}
%{ \displaystyle\mathrm{exp}(Z_{v_{1}^{'}}\cdot Z_{v_{2}^{'}}/\tau)   + \displaystyle\sum_{Z_{v_{n}^{'}} \sim \mathcal{N}} \mathrm{exp}(Z_{v_{1}^{'}}\cdot Z_{v_{n}^{'}}/\tau)},
%\label{eq:3}
%\end{equation}
\vspace{-0.3em}
\section{Generalization on Diving48} 
\vspace{-0.3em}
To further highlight the generalizability of our method to new domains and fine-grained actions, we finetune and evaluate with the challenging Diving48 dataset~\cite{diving48}. %\cite{fame,background_removing,dave2022tclr}. 
It contains 18K trimmed videos for 48 different diving sequences all of which take place in similar backgrounds and need to be distinguished by subtle differences such as the number of somersaults or the starting position. We use standard train/test split and report top-1 accuracy.

In Table~\ref{tab:diving}, we show the performance of our model when pretrained on the full Kinetics-400 and on Mini-Kinetics ($\dagger$). We compare these results to no pretraining, the temporal contrastive baseline pretrained on Kinetics-400, and supervised pretraining on Kinetics-400 with labels.  Our method increases the performance  over training from scratch by 7.9\% and the temporal contrastive baseline by 6.6\%. 
 Our method even outperforms the supervised pretraining baseline by 4.5\%. This suggests that by contrasting tubelets with different motions, our method is able to learn better video representations for fine-grained actions than supervised pretraining on Kinetics.
 When pretraining on Mini-Kinetics (3x smaller than Kinetics-400) the performance of our model does not decrease, again demonstrating the data efficiency of our approach.

\vspace{-0.3em}
\section{Evaluation with R3D and I3D Backbones }
\vspace{-0.3em}
In addition to the R(2+1)-18 backbone,  we also show the performance of our proposed method with other commonly used video encoders \ie, R3D-18~\cite{tran2018closer} and I3D~\cite{carreira2017quo}.
For R3D-18,  we use the same tubelet generation and transformation as that of  R(2+1)D-18, as described in the main paper. For I3D, we change the input resolution to 224x224 and sample the patch size $H^{'}{\times} W^{'}$  uniformly from $[32{\times}32, 128{\times128}]$. For both, we follow the same pretraining  protocol as described in the main paper. 

We compare  with prior works on the standard UCF101~\cite{UCF-101-arxiv} and HMDB51~\cite{HMDB-51-ICCV} datasets. Table~\ref{tab:R3D} shows the results with  Kinetics-400 as the pretraining dataset.   With the I3D backbone, our method outperforms prior works on both UCF101 and HMDB51.
Similarly, with the R3D-18 backbone, we outperform prior works using the RGB modality on UCF101. We also achieve comparable performance to the best-performing method on HMDB51, improving over the next best method by 6.3\%.  
On HMDB51 we also outperform prior works which pretrain on an additional optical flow modality and achieve competitive results with these methods on UCF101.

\begin{table}[t!]
    \centering
    \resizebox{0.6\linewidth}{!}{
    \begin{tabular}{lc}
         \toprule Pretraining &  Top-1\\
         \midrule
         Supervised~\cite{tran2018closer} & 84.5  \\
         \midrule
         None & 81.1 \\
         Temporal Contrast Baseline &  82.4 \\
         \midrule
         \textbf{\textit{This paper}}$^{\dagger}$ & \textbf{89.4} \\
         \textbf{\textit{This paper}}  & \textbf{89.0} \\
        \bottomrule
    \end{tabular}
    }
    \vspace{-0.5em}
        \caption{ \textbf{Generalization on Diving48~\cite{diving48}.} Comparison with temporal contrastive pretraining and supervised pretraining on Diving48. All models use R(2+1)D-18. $\dagger$ indicates pretraining on Mini-Kinetics, otherwise all pretraining was done on Kinetics-400.}
    \label{tab:diving}
    \vspace{-0.5em}
\end{table}

\begin{table}[t!]
\centering
\resizebox{0.75\linewidth}{!}{
    \begin{tabular}{lllcc}
        \toprule
        \textbf{Method} & \textbf{Modality} & UCF & HMDB\\
         \midrule
         \rowcolor{Gray}
         \textbf{I3D} & & & \\
         %\midrule
        SpeedNet~\cite{benaim2020speednet} & RGB &  66.7 & 43.7 \\
        DSM~\cite{wang2021enhancing} & RGB &   74.8 & 52.5 \\
        BE~\cite{background_removing} & RGB & 86.2 & 55.4 \\
        FAME~\cite{fame} & RGB & 88.6 & 61.1 \\
        \textbf{\textit{This paper}}$^{\dagger}$ & RGB&   \textbf{89.5}	& \textbf{64.0} \\
        \textbf{\textit{This paper}} & RGB&   \textbf{89.7}	& \textbf{63.9} \\
         \midrule
         \rowcolor{Gray}
         \textbf{R3D-18} & & & \\
         %\midrule
        %\rowcolor{Gray}
        %$^\dagger$SeLaVi~\cite{selavi}& RGB+Audio & R(2+1)D &   83.1 & 47.1 \\
        %\rowcolor{Gray}
        %$^\dagger$XDC~\cite{alwassel_2020_xdc} & RGB+Audio & R(2+1)D  &  86.8 & 52.6 \\
        %\rowcolor{Gray}
        VideoMoCo~\cite{videomoco-pan2021videomoco} & RGB &  74.1 & 43.6 \\
        RSPNet~\cite{rspnet-chen2020RSPNet} & RGB &   74.3 &41.6 \\
        LSFD~\cite{lsfd} & RGB & 77.2 & 53.7 \\
        MLFO~\cite{mlfo} & RGB &  79.1 & 47.6 \\
        ASCNet~\cite{huang2021ascnet} & RGB &   80.5 & 52.3 \\
        MCN~\cite{mcn} & RGB    & 85.4 & 54.8 \\
        TCLR~\cite{dave2022tclr} & RGB&   85.4 &55.4 \\
        CtP~\cite{ctp-wang2021unsupervised} & RGB&  86.2 & 57.0 \\
        %BE~\cite{background_removing} & RGB & R3D-34 & 87.1 & 55.4\\ 
        TE~\cite{jenni2021time_eqv} & RGB &   87.1 & \textbf{63.6} \\
       \color{gray}
        MSCL~\cite{mscl} & \color{gray} RGB+Flow & \color{gray}   90.7 & \color{gray} 62.3 \\
        \color{gray}
        MaCLR~\cite{xiao2022maclr} & \color{gray} RGB+Flow &  \color{gray} 91.3 & \color{gray} 62.1 \\ %\midrule
%        \rowcolor{visualblue}
        \textbf{\textit{This paper}}$^{\dagger}$ & RGB&   \textbf{88.8}	& {62.0} \\
        \textbf{\textit{This paper}} & RGB&   \textbf{90.1}	& {63.3} \\
        \bottomrule

        \end{tabular}}
      \vspace{-0.5em}    
    \caption{\textbf{Evaluation with I3D and R3D backbones:} on standard  UCF101 and HMDB51 benchmarks. 
    %All models utilize Kinetics-400 for self-supervised pretraining. 
    Gray lines indicate the use of additional modalities during self-supervised pretraining. $\dagger$ indicates pretraining on Mini-Kinetics, otherwise, all models were pretrained on Kinetics-400.
    }
    \label{tab:R3D}
        \end{table}
\begin{table*}[t]
    \centering
    \resizebox{0.87\linewidth}{!}{
    \begin{tabular}{lllrrrr}
    \toprule
    \textbf{Evaluation Factor} & \textbf{Experiment} & \textbf{Dataset} & \textbf{Batch Size} & \textbf{Learning rate} & \textbf{Epochs} & \textbf{Steps} \\ % & \textbf{Eval Metric}\\
    \midrule
    \multirow{2}{*}{\textbf{Standard}} & UCF101 & UCF 101~\cite{UCF-101-arxiv} & 32&  0.0001  & 160 & [60,100,140] \\
 & HMDB51 & HMDB 51~\cite{HMDB-51-ICCV} & 32&  0.0001  & 160 & [60,100,140]  \\
    \midrule
    \multirow{2}{*}{\textbf{Domain Shift}} & SS-v2 & Something-Something~\cite{SS-v2-arxiv} & 32&  0.0001 & 45  & [25, 35, 40] \\
     & Gym-99 & FineGym~\cite{Gym-99-arxiv} & 32&  0.0001 & 160 & [60,100,140] \\
    \midrule
    \multirow{2}{*}{\textbf{Sample Efficiency}}& UCF ($10^3$) & UCF 101~\cite{UCF-101-arxiv} &  32&  0.0001 & 160 & [80,120,140]  \\
     & Gym ($10^3$) & FineGym~\cite{Gym-99-arxiv} & 32&  0.0001 & 160 & [80,120,140]  \\
    \midrule
    \multirow{2}{*}{\textbf{Action Granularity}} & FX-S1 & FineGym~\cite{Gym-99-arxiv} & 32&  0.0001 & 160 & [70,120,140]  \\
     & UB-S1 & FineGym~\cite{Gym-99-arxiv} & 32&  0.0001 & 160 & [70,120,140]  \\
    \midrule
    \multirow{2}{*}{\textbf{Task Shift}} & UCF-RC & UCFRep~\cite{ucfrep-zhang2020context} & 32&  0.00005 & 100 & -  \\
     & Charades & Charades~\cite{charades-sigurdsson:hal-01418216} &  16&  0.0375 & 57 & [41,49]  \\
    \bottomrule
    \end{tabular}}
    \vspace{-0.7em}
    \caption{\textbf{Training Details} of finetuning on various downstream datasets and tasks.}
    \label{tab:finetuning}
        \vspace{-0.8em}
\end{table*}

\vspace{-0.3em}
\section{Evaluation on Kinetics Dataset}
\vspace{-0.3em}
To show whether our tubelet-contrastive pretraining can improve the performance of downstream tasks when plenty of labeled data is available for finetuning, we evaluate it on the Kinetics-400~\cite{Kinetics-400-arxiv} dataset for the task of action classification. The dataset contains about 220K labeled videos for training and 18K videos for validation. As evident from Table~\ref{tab:k400},  such large-scale datasets can still benefit from our pretraining with a 3.4\% improvement over training from scratch and 0.7\% over the temporal contrast baseline.

\begin{table}[t!]
    \centering
    \resizebox{0.6\linewidth}{!}{
    \begin{tabular}{lc}
         \toprule Pretraining &  Top-1\\
         \midrule
         None & 61.4 \\
         Temporal Contrast Baseline &  64.1 \\
         \midrule
         \textbf{\textit{This paper}}  & \textbf{64.8} \\
        \bottomrule
    \end{tabular}
    }
    \vspace{-0.5em}
    \caption{\textbf{Kinetics-400 Evaluation.} Comparison with temporal contrastive pretraining for large-scale action recognition. All models use R(2+1)D-18 and pretraining was done on Kinetics-400 training set.}
    \label{tab:k400}
    \vspace{-0.7em}
\end{table}

\vspace{-0.3em}
\section{Finetuning Details} 
\vspace{-0.3em}
During finetuning, we follow the setup from the SEVERE benchmark~\cite{thoker2022severe} which is detailed here for completeness. For all tasks, we replace the projection of the pretrained model with a task-dependent head. \\
\noindent\textbf{Action Recognition}. 
Downstream settings which examine domain shift, sample efficiency, and action granularity all perform action recognition. We use a similar finetuning process for all experiments on these three factors. During the training process, a random clip of 32 frames is taken from each video and standard augmentations are applied: a multi-scale crop of 112x112 size, horizontal flipping, and color jittering. The Adam optimizer is used for training, with the  learning rate, scheduling, and total number of epochs for each experiment shown in Table~\ref{tab:finetuning}. During inference, 10 linearly spaced clips of 32 frames each are used, with a center crop of 112x112. To determine the action class prediction for a video, the predictions from each clip are averaged. For domain shift and sample efficiency, we report the top-1 accuracy.
 For action granularity experiments we report mean class accuracy, which we obtain by computing accuracy per action class and averaging over all action classes. \\
\noindent\textbf{Repetition counting}. 
The  implementation follows the original repetition counting work proposed in  UCFrep work~\cite{ucfrep-zhang2020context}.
From the annotated videos, 2M sequences of 32 frames with spatial size 112x112 are constructed. These are  used as the input. The model is trained with a batch size of 32 for 100 epochs using the Adam optimizer with a learning rate of 0.00005. For testing, we report mean counting error following\cite{ucfrep-zhang2020context}. \\
\noindent\textbf{Multi-label classification on Charades}. Following \cite{large-scale-feichtenhofer2021large},
 a per-class sigmoid output is utilized for multi-class prediction. During the training process, 32 frames are sampled with a stride of 8. %Since this task requires a longer temporal context, using more frames with a higher sampling rate can improve the results. 
Frames are cropped to 112x112 and random short-side scaling, random spatial crop, and horizontal flip augmentations are applied. The model is trained for a total of 57 epochs with a batch size of 16 and a learning rate of 0.0375. A multi-step scheduler with $\gamma = 0.1$ is applied at epochs [41, 49]. During the testing phase,  spatiotemporal max-pooling is performed over 10 clips for a single video. We report mean average precision (mAP) across all classes.    

\noindent\textbf{SSv2-Sub details}. We use a subset of Something-Something v2 for ablations. In particular, we randomly sample 25\% of the data from the whole train set and spanning all categories. This results in a subset consisting of 34409 training samples from 174 classes. We use the full validation set of Something-Something v2 for testing.

\begin{table}[t!]
    \centering
    \resizebox{0.8\linewidth}{!}{
    \begin{tabular}{lcc} \toprule Transformation & UCF ($10^{3}$) & Gym ($10^{3}$)  \\
         \midrule
         None   &                           63.0 & 45.6      \\
         \midrule
         Scale  &                           65.1 & 46.5      \\
         Shear  &                           65.2 & 47.5      \\
         Rotate &                           65.5 & 48.0      \\
         \midrule  
         Scale + Shear  &                   65.2 & 46.0      \\
         Rotate + Scale   &                 65.4 & 46.9      \\
         Rotate + Shear   &                 65.3 & 45.7      \\
         Rotate + Scale + Shear  &          65.6 & 46.0      \\
        \bottomrule
    \end{tabular}}
      \vspace{-0.8em}
    \caption{\textbf{Tubelet Transformation Combinations.} Combining transformations doesn't give a further increase in performance compared to using individual transformations. }
    \label{tab:ablation_transformations_combination}
    \vspace{-0.5em}
\end{table}

\vspace{-0.5em}
\section{Tubelet Transformation Hyperparameters} 
\vspace{-0.3em}
Table~\ref{tab:ablation_transformations_combination} shows the results when applying multiple tubelet transformations in the tubelet generation. While applying individual transformations improves results, combing multiple transformations doesn't improve the performance further. This is likely because rotation motions are common in the downstream datasets while scaling and shearing are less common. %such transformations result in simulating motion with realistic transformations which do not resemble the motion of objects/persons in downstream videos. 

Table~\ref{tab:ablation_transformations_range} shows an ablation over $\mathrm{Min}$ and $\mathrm{Max}$ values for tubelet transformations.  In the main paper, we use scale values between 0.5 and 1.5, shear values between -1.0 and 1.0, and rotation values between -90 and 90. Here, we experiment with values that result in more subtle and extreme variations of these transformations.  We observe that all values for each of the transformations improve over no transformation. Our model is reasonably robust to these choices in hyperparameters, but subtle variations \eg,  scale change between 0.5 to 1.25 or shear from 0.75 to 0.75 tend to be slightly less effective. %We also observe that performance can be improved be extending the  range of these hyperparameters \eg rotation between -180 to 180.

\begin{table}[t!]
    \centering
    \resizebox{0.65\linewidth}{!}{
    \begin{tabular}{llcc} \toprule Min & Max & UCF ($10^{3}$) & Gym ($10^{3}$)  \\
         \midrule
         %\hdashline
         \rowcolor{Gray}
         \textbf{None} & & & \\
         - & - &                            63.0 & 45.6 \\
         \rowcolor{Gray}
         \textbf{Scale}  &  & &\\
         0.5 & 1.25  &                  65.6 & 45.3      \\
         0.5 & 1.5  &                   65.1 & 46.5      \\
         0.5 & 2.0  &                   65.6 & 46.0      \\
         %\midrule
         \rowcolor{Gray}
         \textbf{Shear}  &  & &\\
         -0.75 & 0.75  &                   64.4 &   47.5    \\
         -1.0 & 1.0  &                   65.2 & 48.0      \\
         -1.5 & 1.5  &                     65.2 &   47.5      \\
         %\midrule
         \rowcolor{Gray}
         \textbf{Rotation}  &  & &\\
         -45 & 45                        &    65.2 & 49.3      \\
         -90 & 90                       &    65.5 & 48.0      \\
         -180 & 180                        &    65.6 & 49.6     \\  
        \bottomrule
    \end{tabular}}
      \vspace{-0.85em}
    \caption{\textbf{Tubelet Transformation Hyperparameters.} We change $\mathrm{Min}$ and $\mathrm{Max}$ values for tubelet transformations. Our model is robust to changes in these parameters, with all choices tested giving an improvement over no tubelet transformation.}
    \label{tab:ablation_transformations_range}
\end{table}

\vspace{-0.8em}
\section{Tubelets vs. Randomly Scaled Crops} 
\vspace{-0.5em}
To show that our proposed tubelets inject useful motions in the training pipeline, we compare them with  randomly scaled crops.  In particular, we randomly crop, scale, and jitter the patches pasted into the video clips when generating positive pairs and pretrain this and our model on Mini-Kinetics.  Table~\ref{tab:scaledcrops} shows that our proposed motion tubelets outperform such  randomly scaled crops in all downstream settings. This validates that the spatiotemporal continuity in motion tubelets is important to simulate smooth motions for learning better video representations. 
\begin{table}[t]
    \centering
    \resizebox{\linewidth}{!}{\begin{tabular}
     {l  cccc}
    \toprule
      %\addlinespace[0.1cm]
         &   \multicolumn{1}{c}{UCF ($10^3$)} & \multicolumn{1}{c}{Gym ($10^3$)}  & \multicolumn{1}{c}{SSv2-Sub} & \multicolumn{1}{c}{UB-S1}  \\ 
         \midrule
         Randomly Scaled Crops & 59.5 & 37.5 & 44.8 & 87.0      \\
         Tubelets & 65.5 & 48.0 &47.9 &90.9       \\
         \bottomrule
    \end{tabular}
    }
    \vspace{-0.85em}
    \caption{\textbf{Tubelets vs Randomly Scaled Crops}. Our tubelets generate smooth motions to learn better video representations than strongly jittered crops.}
\label{tab:scaledcrops}
\end{table}

\vspace{-0.9em}
\section{Per-Class Results}
\begin{figure}[t!]
\centering
\includegraphics[width=0.95\linewidth]{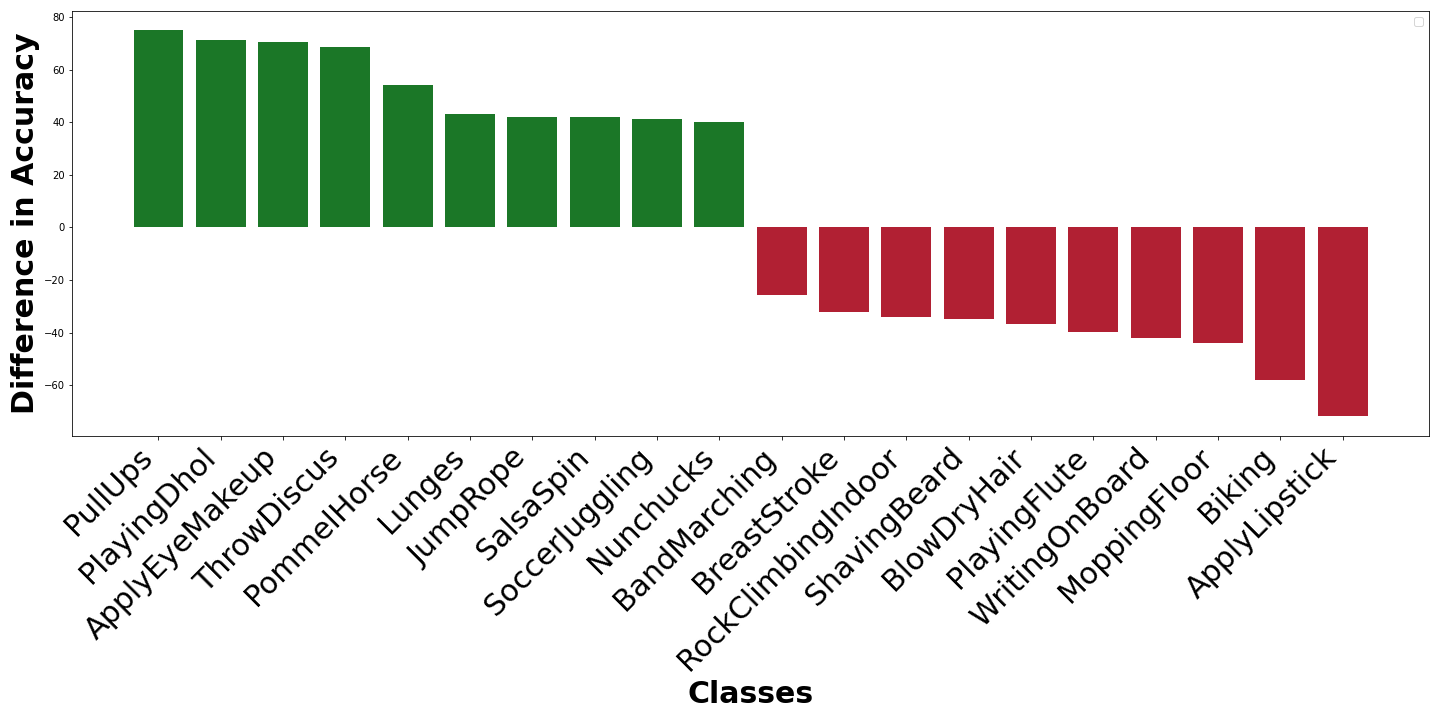}
\vspace{-1.0em}
\caption{\textbf{Per-Class Accuracy Difference} on UCF ($10^3$) between our model and the temporal contrastive baseline. We show the 10 actions with the highest increase and decrease. Our model can better distinguish classes requiring motion but loses some ability to distinguish spatial classes.}
\label{fig:ucf_bar}
\end{figure}

\vspace{-0.3em}
Examining the improvement for individual classes gives us some insight into our model. Figure~\ref{fig:ucf_bar} shows the difference between our approach and the baseline for the 10 classes in UCF ($10^3$) with the highest increase and decrease in accuracy.  Many of the actions that increase in accuracy are motion-focused, \eg, pullups, lunges and jump rope. Other actions are confused by the baseline because of the similar background, \eg, throw discus is confused with hammer throw and apply eye makeup is confused with haircut. The motion-focused features our model introduces help distinguish these classes. However, our model does lose some useful spatial features for distinguishing classes such as band marching and biking.

%\vspace{-5pt}
\vspace{-0.9em}
\section{Class Agnostic Activation Maps} 
\begin{figure*}[t!]
\centering
\includegraphics[width=0.8\linewidth]{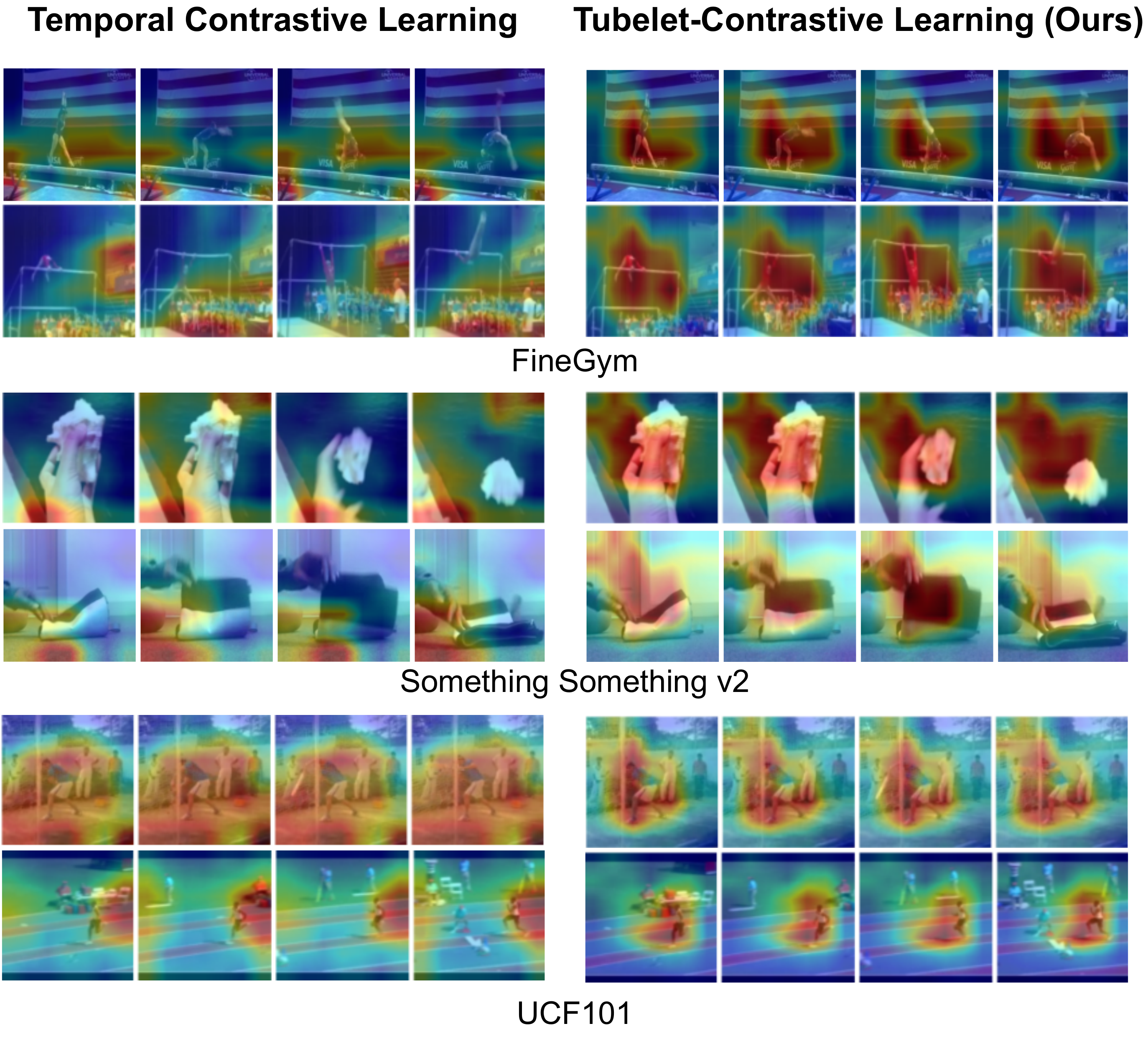}
\caption{\textbf{Class-Agnostic Activation Maps Without Finetuning} for the temporal contrastive baseline and our tubelet contrast for different downstream datasets.  Our model better attends to regions with motion irrespective of the domain. }
\label{fig:caam_supp}
\end{figure*}

\vspace{-0.3em}
Figure~\ref{fig:caam_supp} show more examples of class agnostic activation maps~\cite{CAAM} for video clips from various downstream datasets. Note that no finetuning is performed, we directly apply the representation from our tubelet contrastive learning pretrained on Kinetics-400. For examples from FineGym, Something Something v2, and UCF101, we observe that our approach attends to regions with motion while the temporal contrastive baseline mostly attends to background. 

\vspace{-0.9em}
\section{Limitations and Future Work}
\vspace{-0.5em}
There are several open avenues for future work based on the limitations of this work. First, while we compare to transformer-based approaches, we do not present the results of our tubelet-contrast with a transformer backbone. Our initial experiments with a transformer-based encoder~\cite {dosovitskiy2020image} did not converge with off-the-shelf settings.  We hope future work can address this problem for an encoder-independent solution.
Additionally, we simulate tubelets with random image crops that can come from both background and foreground regions. Explicitly generating tubelets from foreground regions or pre-defined objects is a potential future direction worth investigating. Finally, we only simulate tubelets over short clips, it is also worth investigating whether long-range tubelets can be used for tasks that require long-range motion understanding.

